\documentclass[journal]{IEEEtran}

\ifCLASSOPTIONcompsoc
\usepackage[nocompress]{cite}
\else
\usepackage{cite}
\fi

\ifCLASSINFOpdf
\usepackage[pdftex]{graphicx}
\graphicspath{{../pdf/}{../jpeg/}}
\DeclareGraphicsExtensions{.pdf,.jpeg,.png}
\else
\usepackage[dvips]{graphicx}
\graphicspath{{../eps/}}
\DeclareGraphicsExtensions{.eps}
\fi

\usepackage{graphicx}
\usepackage{subfigure}
\usepackage{amsmath}
\usepackage{algorithm}
\usepackage{algorithmicx}
\usepackage{algpseudocode}
\usepackage{multirow}
\usepackage{amsfonts}
\usepackage{float}
\usepackage{color}
\usepackage{amsfonts}
\usepackage{soul}
\usepackage{textcomp}

 \usepackage{dblfloatfix}
\usepackage{url}

\begin{document}
%
\title{Clairvoyance: Intelligent Route Planning \\ for Electric Buses Based on Urban Big Data}
%
%
%

\author{Xiangyong~Lu,~\IEEEmembership{}
	Kaoru Ota,~\IEEEmembership{Member,~IEEE,}
	Mianxiong Dong,~\IEEEmembership{Member,~IEEE,}
	Chen Yu,~\IEEEmembership{Member,~IEEE,}
	and~Hai~Jin,~\IEEEmembership{Fellow,~IEEE}
	\IEEEcompsocitemizethanks{
		\IEEEcompsocthanksitem X. Lu is with the Graduate School of Information Sciences, Tohoku University, Sendai, Japan. 
		\protect\\
		E-mail:xylu@vision.is.tohoku.ac.jp;xiangyonglu@126.com;
		\IEEEcompsocthanksitem C. Yu and H. Jin are with the 
		Serviecs Computing Technology and System Lab,
		Big Data Technology and System Lab,
		Cluster and Grid Computing Lab,
		School of Computer Science and Technology,
		Huazhong University of Science and Technology, Wuhan, 430074, China.
		\protect\\
		E-mail: yuchen@hust.edu.cn; hjin@hust.edu.cn
		\IEEEcompsocthanksitem K. Ota, M. Dong are with the Department of Information and Electronic Engineering, Muroran Institute of Technology, Muroran, Hokkaido, Japan.
		\protect\\
		E-mail: ota@csse.muroran-it.ac.jp; mx.dong@csse.muroran-it.ac.jp}
	\thanks{}}

%
%

\markboth{ IEEE TRANSACTIONS ON INTELLIGENT TRANSPORTATION SYSTEMS ,~Vol.~, No.~, ~2020}%
{Shell \MakeLowercase{\textit{et al.}}: Bare Demo of IEEEtran.cls for IEEE Journals}
%



\maketitle

\begin{abstract}
Nowadays many cities around the world have introduced electric buses to optimize urban traffic and reduce the local carbon emissions. In order to cut the carbon emissions and maximize the utility of electric buses, it is important to choose suitable routes for them. Traditionally, route selection is on the basis of dedicated surveys, which are costly in time and labor. In this paper, we mainly focus attention on planning electric bus routes intelligently, depending on the unique needs of each region throughout the city. We propose \emph{Clairvoyance}, a route planning system that leverages a deep neural network and a multilayer perceptron to predict the future people's trips and the future transportation carbon emission in the whole city, respectively. Given the future information of people's trips and transportation carbon emission, we utilize a greedy mechanism to recommend bus routes for electric buses that will depart in an ideal state. Furthermore, representative features of the two neural networks are extracted from the heterogeneous urban datasets. We evaluate our approach through extensive experiments on the real-world data sources in Zhuhai, China. The results show that our designed neural network-based algorithms are consistently superior to the typical baselines. Additionally, the recommended routes for electric buses are helpful in reducing the peak value of carbon emissions and making full use of electric buses in the city. 
\end{abstract}


%
\IEEEpeerreviewmaketitle

\section{Introduction}
%
%
%
%
\IEEEPARstart{E}{lectric} buses are the significant contributors to reduce transportation carbon emission and energy consumption. Besides, the cost savings from fuel more than offset the cost of electric bus and infrastructure over the lifetime of a bus. In recent years, a growing number of cities have taken steps to introduce electric buses, in order to optimize urban transport structure and reduce carbon emissions and fuel consumption. However, the number of electric buses introduced in cities is constantly limited. Meanwhile, given the large-scale infrastructure investment necessary to support electric bus transportation, it is essential for policy makers to maximize the utility of electric buses. A key factor in making full use of electric buses is to choose the proper routes for them according to the local conditions.

Due to the unique geographical, environmental and cultural characteristics in each region of the city, transportation carbon emission and people's trips vary drastically with place and time.
When a limited number of electric buses are allocated the fixed routes, it is difficult to meet the dynamic demand of people's trips and reducing carbon emissions. To meet the diverse demand of the city, routes should be assigned dynamically for electric buses. In this paper,we want to answer the practical question: When a limited amount of electric buses are introduced in a city, which route should be dynamically selected for each electric bus that will depart in an ideal state? We aim to predict the potential transportation carbon emission and the potential number of people's bus trips throughout the city to guide the electric bus route selection. The task is extraordinary challenging due to the following three reasons: 

\emph{First}, transportation carbon emission is affected by multiple factors, such as human mobility, traffic flow, and meteorology. The factors differ in different regions and change over time. For example, traffic flows on the main roads are often larger than those of ordinary roads during peak commuting hours. The changes of these factors contribute to the difference in transportation carbon emission between all regions of the city. Consequently, we should identify the main factors that influence the transportation carbon emission and extract the efficient features from the heterogeneous data sources in each region.

\begin{figure}[ht]
		 \centering 
	\subfigure[Eight random bus routes in Zhuhai. ]{
		\label{pic:eightBusRoutes} 
		\includegraphics[width=1.45in]{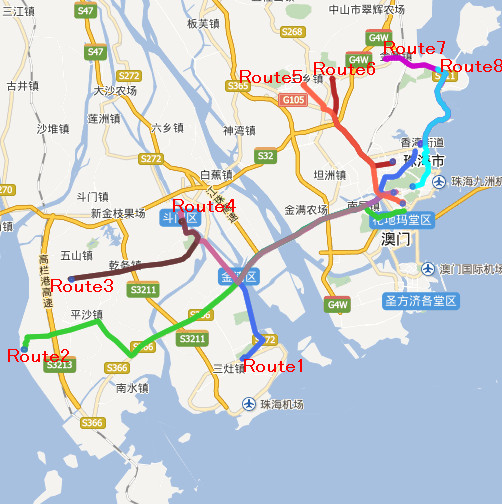}}
	\subfigure[The average number of people's trips on the eight routes in one hour after 11 a.m. (Sept. 1 to Sept. 15, 2015).]{
		\label{pic:NumOfBusPass}
		\includegraphics[width=1.45in]{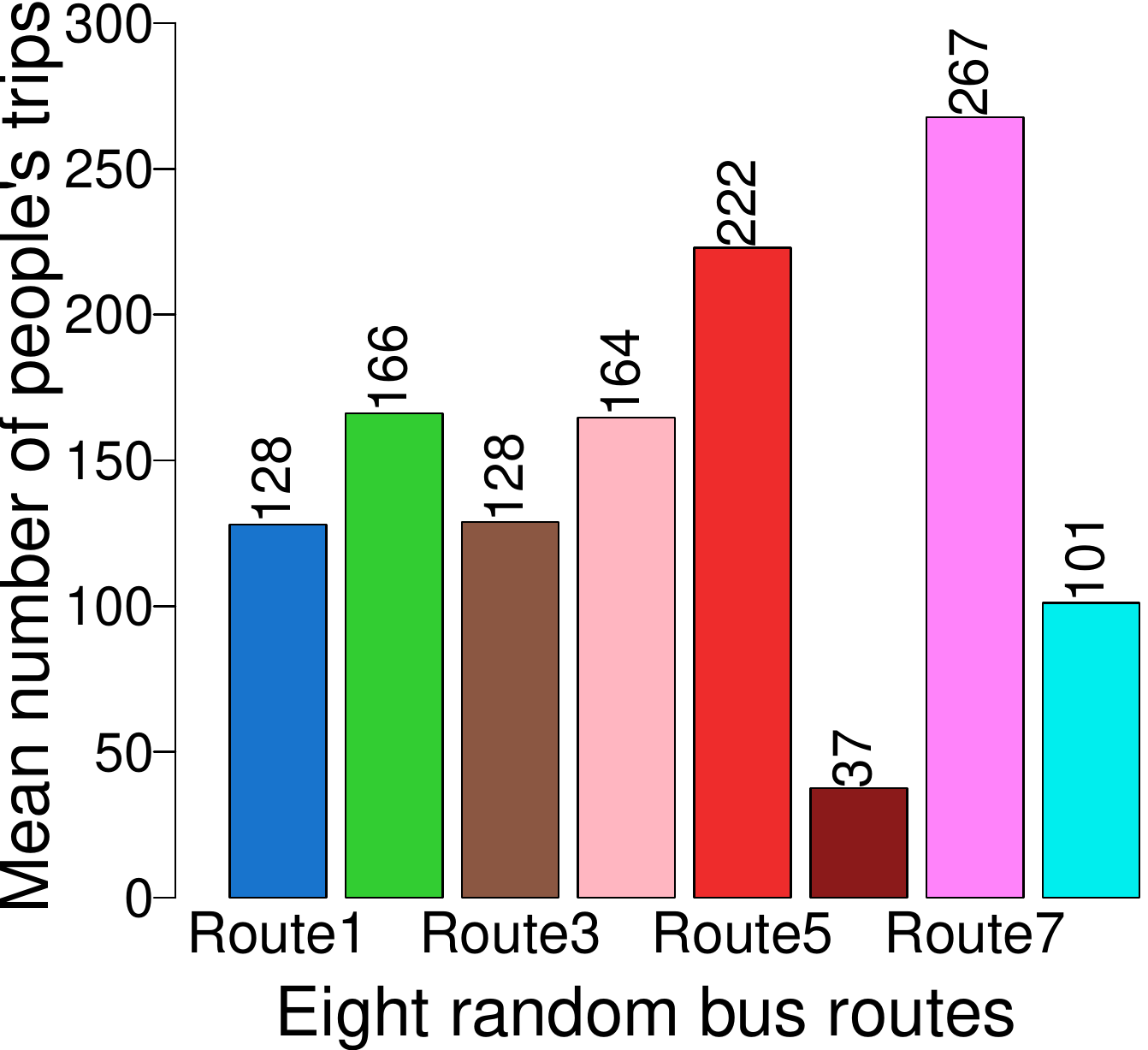}}
	\hspace{0in}
	\subfigure[ The number of people's trips on the five routes in one hour after 11 a.m. each day in September 2015.
	]{
		\label{pic:BTDVaries30Days11oclock} 
		\includegraphics[width=1.6in]{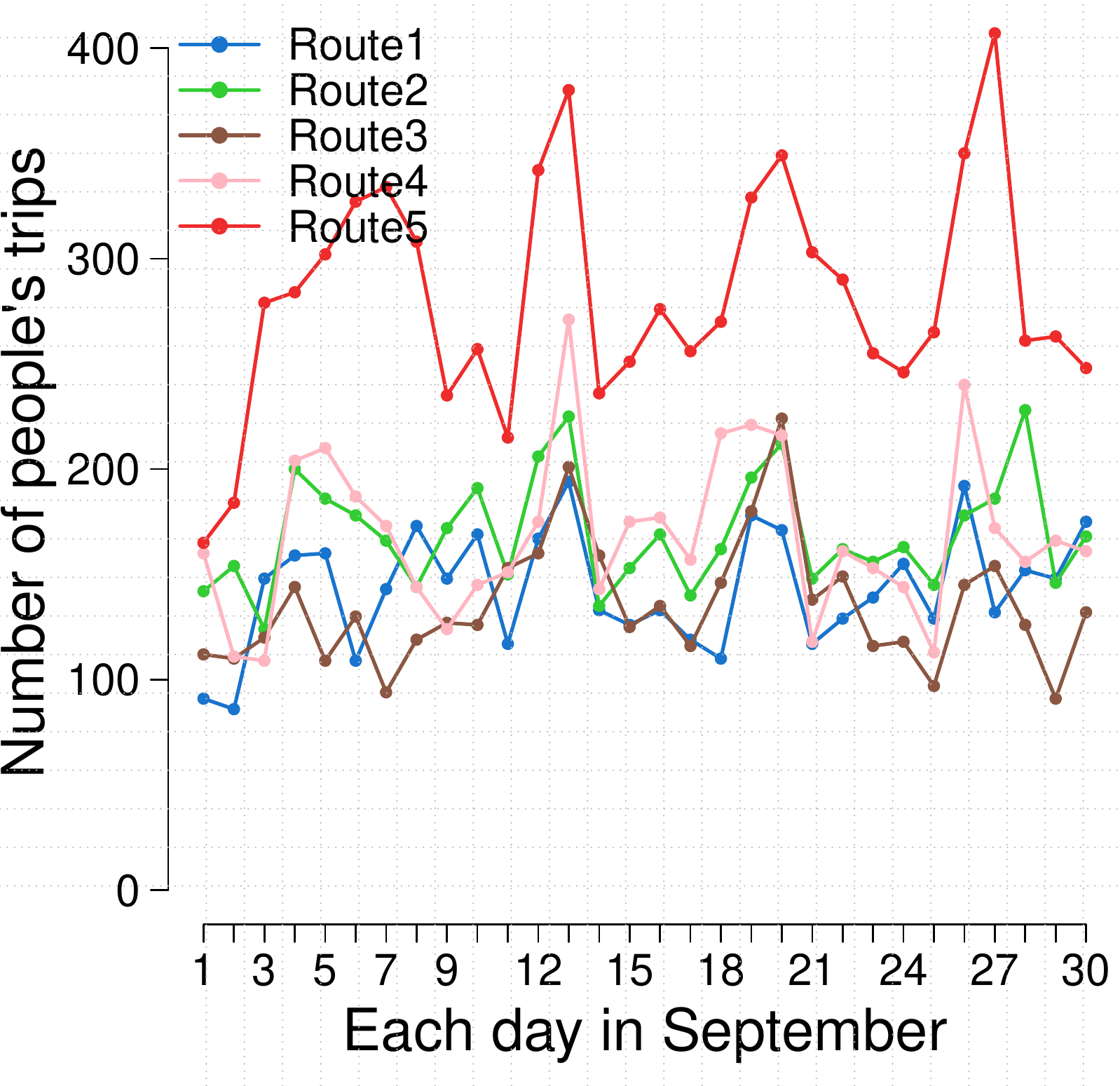}}
	\hspace{0in}
	\subfigure[The number of people's trips on the five routes in each hour of September 5, 2015.] {
		\label{pic:linkPassPerHR}
		\includegraphics[width=1.6in]{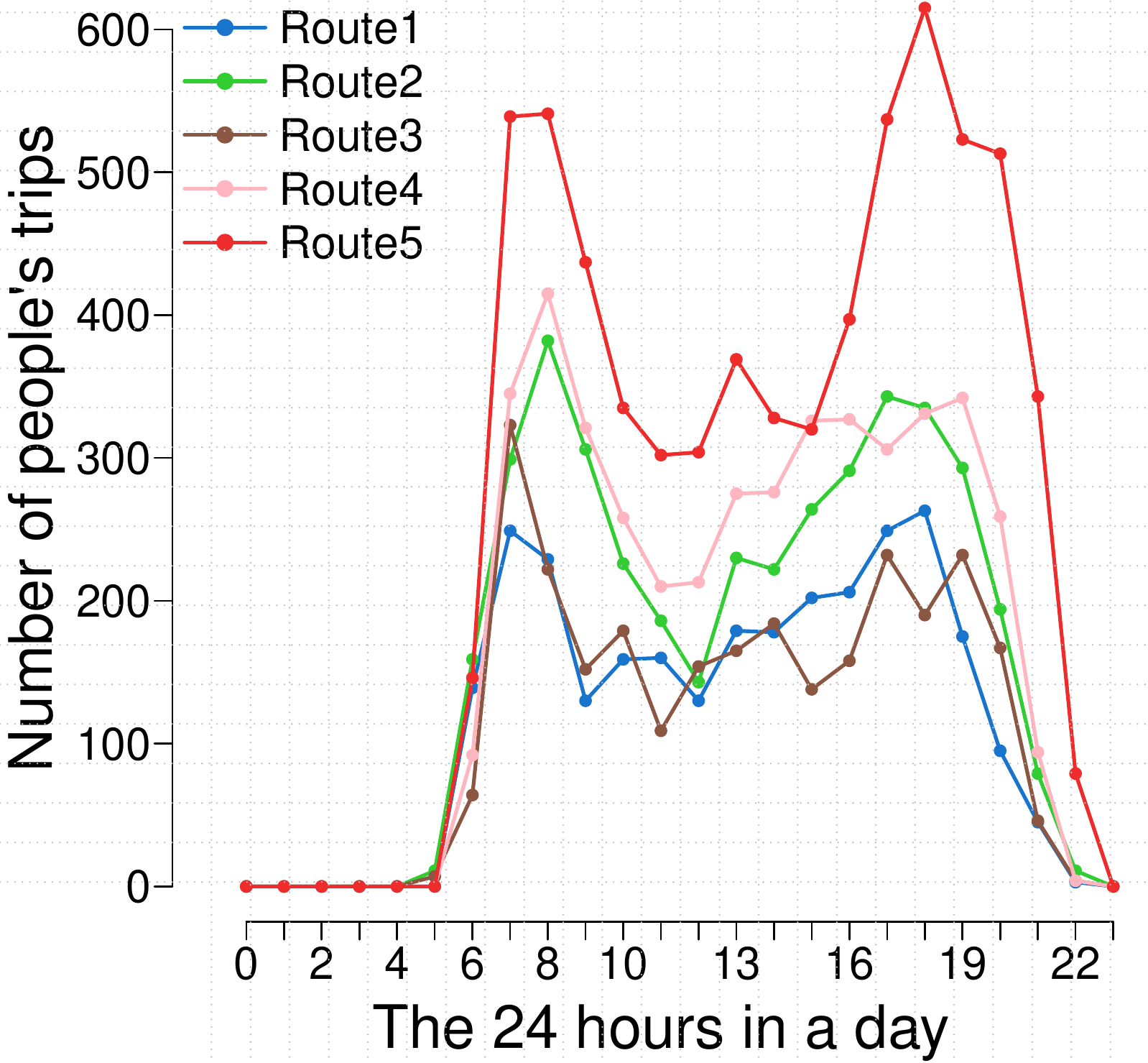}}
	
	\caption{Changes in the number of people's trips on the bus routes in Zhuhai.} 
	\label{pic:urbanLayout} 
\end{figure}

\emph{Second}, on each road in the city, the number of people's bus trips significantly changes over location and time. As shown in Fig. \ref{pic:eightBusRoutes}, eight routes are randomly selected from all bus routes in Zhuhai. Each bus route locates in different areas of the city. Fig. \ref{pic:NumOfBusPass} outlines the average numbers of people's trips on the eight routes in one hour after 11 a.m. during the period from Sept. 1 to Sept. 15, 2015. 
Meanwhile, as described in Fig. \ref{pic:BTDVaries30Days11oclock} and \ref{pic:linkPassPerHR}, the number of people's trips on the first five routes are diverse and vary with time tremendously. 
For example, there is a dramatically increase in the amount of trips during peak commuting hours, such as between 8 and 9 a.m. and between 6 and 7 p.m.. Accordingly, to obtain the future number of people's trips on all bus routes, we should dig out the unique spatial-temporal characteristics of each route.

\emph{Third}, traditionally policy makers obtain transportation carbon emission information, by calculating the combustion of fossil fuel in the whole transportation sector \cite{intergovernmental20062006}. The information of people's bus trips is collected by dedicated surveys. However, these two methods are almost impossible to obtain the real-time and accurate information. Furthermore, affected by unforeseen emergencies, the transportation carbon emission and people's trips in each area change sharply. As numerous emergencies are noticeably rare, a general statistical model will be difficult to predict the sudden changes.

To address the aforementioned challenges, in this paper, we divide the whole city into disjoint and uniform grids. Subsequently, we predict transportation carbon emission in each grid, based on the taxi \emph{GPS} data, urban road data, points of interests and meteorological data. Meanwhile, around each bus station on the routes, we define a uniform affecting region, as shown in Fig. \ref{pic:affectingRegion}. 
The future number of people's trips on each bus route is then inferred, based on the extracted features in the affecting regions.

In our method, we propose a neural network-based model that consists of two separated predictors. One is a carbon emission predictor based on a three-layer perceptron neural network. The other is trip demand predictor based on a deep neural network. Given the multiple spatio-temporal urban data, we train our model to predict the transportation carbon emission in each grid and the number of bus trips on each bus route. Combining the results of these two predictors, we devise a greedy mechanism to recommend suitable bus routes for electric buses that will depart. In this paper, our contribution lies in the following three aspects:
\begin{itemize}
	\item We propose a route planning system \emph{Clairvoyance}, which uses a deep neural network to predict the future number of bus trips on each bus route and utilizes a three-layer perceptron to forecast the future transportation carbon emission in each grid throughout a city. Furthermore, based on these two prediction results, we devise a greedy mechanism to recommend suitable bus routes for each electric bus that will depart.
	\item We extract efficient representative features from the heterogeneous urban data sources for our learning methods. Moreover, our mining features not only contribute to our application, but also contribute to the general problems of sensing urban dynamics.
	\item We evaluate our approach utilizing the multiple datasets in Zhuhai, China, that is meteorological data, urban road data, points of interest data, bus \emph{IC} card data, bus \emph{GPS} trajectories generated by over 1,500 buses from July 1st to October 14th, 2015 and the taxi \emph{GPS} trajectories generated by over 3,000 taxicabs from July 1st to October 14th, 2015. We justify the advantages of our designed neural network-based algorithms over several baselines.
\end{itemize}

The rest of this paper is organized as follows.  In section II, we introduce the works related to our research. Preliminary definitions and the framework of our system are described in section III. Feature extraction are listed in section IV. Feature learning and inference are discussed in section V. Experimental evaluation is shown in section VI. Finally, we conclude our work in section VII.

\section{Related Work}
%
%

\subsection{Urban transit planning}

In recent years, many cities have invested heavily in urban traffic construction and transit planning.
Transit planning and optimization are mainly focused on two aspects, that is cost and service \cite{litman2015evalbenefitsCosts}.
Through the effective transit planning, the urban transit systems will serve a greater number of people and obtain a high level of customer satisfaction.
In recent decades, urban transit system has benefited hugely from the increasing number of sensors deployed in roads, vehicles and buildings. 
Moreover, based on the innovative data processing \cite{quan2017efficient} \cite{zikopoulos2011bigdata} and mining technologies  \cite{witten2016MLtechniques} \cite{zheng2015trajectory}, a series of studies on urban transportation have appeared, such as transit planning \cite{oliveira2019pathplanning} \cite{osang2019livetravelroutes}, vehicle scheduling \cite{zhang2020fuzzytimetable} \cite{achar2019bustimetable}, and traffic flow\cite{treiber2013traffic} \cite{abadi2015traffic}. 

For public transit frequency setting, Hadas \cite{hadas2012public} proposes a cost-based approach, which contains two main cost elements: empty-seat driven and overload and unserved demand. Through minimizing the total cost incurred in connection with either frequency or vehicle capacity, his method can dramatically improve the accuracy
in the frequency setting.
For traffic flow prediction, Lv \cite{lv2015trafficFlow} devises a stacked autoencoder model, utilizing the datasets collected from the Caltrans Performance Measurement System (PeMS) database. While Zheng \cite{zheng2019deeplearningtrafficflow} designs a deep and embedding learning approach to improve the prediction accuracy of traffic flow, through explicitly extracting fine-grained information from massive traffic data, route structure, and weather conditions.
Moreover, in order to solve the load scheduling problem of plug-in hybrid electric vehicles, Kang \cite{kang2017scheduling} proposes a novel weight aggregation multi-objective particle swarm optimization (WA-MOPSO), which is based on the existing multi-objective evolutionary algorithms, such as non-dominated sorting genetic algorithm-II. 
Based on the real case of bus operation in Shenzhen, China, Jiang \cite{jiang2018scheduling} solves the scheduling problem of the regular charging electric buses by a neighborhood search heuristic.
For the optimal dispatching of electric and hybrid buses, Rinaldi \cite{rinaldi2018dispatching} introduces a Mixed Integer Linear Program, which objective function is of minimizing the total operational cost. 
In order to collaboratively optimize the vehicle scheduling and charging scheduling of the mixed bus fleet, Zhou \cite{zhou2020collaborative} develops a Multi-objective Bi-level programming model. For the upper-level vehicle scheduling problem, he designs an improved iterative neighborhood search algorithm to minimize the operating cost and carbon emissions. For the lower-level charging scheduling problem, he proposes a greedy search strategy to minimize the charging cost. Compared with the previous works, we devise a neural network-based system to infer suitable routes for electric buses that will depart, based on urban big data. Two main parts in our system are to predict the future number of bus trips on each route and transportation carbon emission in each region throughout the city.

\subsection{Urban Computing}
Urban computing is an interdisciplinary field, where computer sciences are used in urban areas, contributing to the improvement of urban environment, human life quality, and city operation systems \cite{zheng2014urban}. In recent decades, the rapid progress of urbanization and industrialization has caused lots of urban issues, such as traffic jams, energy consumption, environmental pollution. In order to tackle these issues, nowadays numerous research institutions around the world begin to do in-depth research in the area of urban computing. 
Based on the air quality data, meteorological data and weather forecasts, Zheng \cite{zheng2015forecastingAir} proposes a hybrid predictive model to forecasting fine-grained air quality. This model consists of four major components, that is a linear regression-based temporal predictor, a neural network-based spatial predictor, a dynamic aggregator, and an inflection predictor. To predict the air quality, he also employs a semi-supervised learning approach based on a co-training framework, which comprises two separated classifiers: an artificial neural network and a linear-chain conditional random field \cite{zheng2013u_air}.

Moreover,  Shang \cite{shang2014gasConsumption} proposes an unsupervised Bayesian Network model and a context-aware matrix factorization approach to infer separately the traffic volume and travel speed on each road segment. Given the resulting traffic volume and travel speed, he calculates the gas consumption and emissions.
Chen \cite{chen2015bikeStation} formulates the bike station placement issue as a bike trip demand prediction problem. He predicts the potential bike trip demand, utilizing the linear regression-and-ranking and artificial neural network regression-and-ranking.
Karamshuk \cite{karamshuk2013retailStore} puts forward a supervised learning model and a supervised regression model to optimize the placement of retail stores, using the feature sets extracted by studying the predictive power of features on the popularity of retail stores.
In order to deal with the non-linear effects of various external factors on urban dynamics, Shimosaka \cite{shimosaka2015forecasting} proposes a low-rank bilinear Poisson regression model to analyze and predict the urban dynamics. 
Based on the large user activity data in the location based social networks, Yang \cite{yang2015modeling} presents a context-aware fusion framework consisting of the spatial-temporal activity preference models to infer the user activity preference. 
Compared with their works, we devise a greedy mechanism to recommend routes for electric buses, based on a deep neural network pre-trained by stacking denoising autoencoders and a multi-layer perceptron neural network. 

\section{Overview}
\subsection{Preliminary}

\noindent\emph{\textbf{Definition 1: }} \emph{Bus Trip Demand.} It is the number of people's bus trips in an area during a specified time period in the future. For a specified time period $T$, the bus trip demand $Q_r$  on the bus route $r$ is the sum of bus trips $q$ generated at the bus station $s$ on the $r$, as formulated in Equation \ref{eq:BusTripDemand}. 
\begin{equation}
\label{eq:BusTripDemand}
Q_{r} = \sum_{i\in \mathbb{Z}^{+}} q_{s_{i}}, \quad s_{i} \in S_{r}
\end{equation}
where $q_{s_{i}}$ is the bus trips generated at the the bus station $s_{i}$ with index $i$.  $S_{r}$ represents the set of bus stations on the $r$.

\noindent\emph{\textbf{Definition 2: }} \emph{Transportation Carbon Emission.} It is a measure of the total amount of carbon dioxide emissions, which are directly or indirectly caused by the traffic. 

\noindent\emph{\textbf{Definition 3:}} \emph{Top-Down Approach.} It is a calculation method of transportation carbon emission, which is given by the 2006 $IPCC$ guidelines \cite{intergovernmental20062006}. This approach is formulated as:
\begin{equation}
\label{eq:topDown}
W = \sum_{i\in \mathbb{N}}\sum_{j\in \mathbb{N}} K_{i,j} \cdot n_{i,j} \cdot l_{i,j} \cdot e_{i,j}
\end{equation}
where $W$ is the total amount of transportation carbon emission. $i$ is a kind of vehicle types. $j$ is a type of fossil fuel. $K_{i,j}$ is the carbon dioxide emission coefficient of $j$ type fossil fuel, which is consumed by the $i$ type vehicles.  $n_{i,j}$ and $l_{i,j}$ represent the number and transport mileage of the $i$ type vehicles, which consume the $j$ type fossil fuel.  $e_{i,j}$ is the carbon emission intensity per $j$ type fuel and for the $i$ type vehicles.

\noindent\emph{\textbf{Definition 4: }}\emph{Grid.} A grid is an area of the city (e.g.,5$km$ $\times$ 5$km$) \cite{lu2017predicting}. The location of each grid is the geographic center of the grid. In the experiments, we divide the Zhuhai city into disjoint and uniform grids, and each grid $g$ has the unique amount of transportation carbon emission $g.W$. 

\begin{figure}[ht]
	\centering 
	\subfigure[Geographical distribution of affecting regions on the bus routes in Zhuhai.]{
		\label{pic:affectRegion} 
		\includegraphics[width=1.6in]{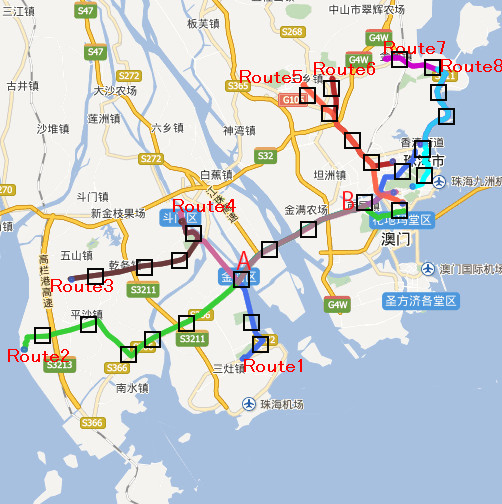}} 
	\hspace{0.2in} 
	\subfigure[The affecting region $A$ and $B$ on the bus routes in Zhuhai.] {
		\label{pic:region}
		\includegraphics[height=1.6in]{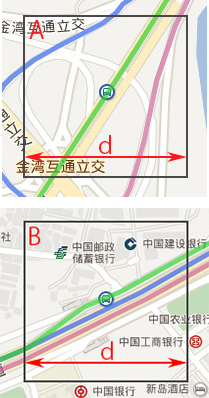}} 
	\caption{Affecting regions on bus routes in Zhuhai city.}
	\label{pic:affectingRegion}
\end{figure}

\noindent\emph{\textbf{Definition 5: }} \emph{Affecting Region.}  An affecting region $\gamma$ is an area around the bus station. The location of each region $\gamma.loc$ is the geographic coordinate of corresponding bus station. In the experiments, each affecting region is a square area with side length 0.5$km$ ($d=0.5km$) centered on the bus station, as shown in Fig. \ref{pic:affectingRegion}.  On each bus route, we randomly select fifteen affecting regions.

\noindent\emph{\textbf{Definition 6: }} \emph{Trajectory.} A spatial trajectory is a sequence of time-ordered spatial points, $\tau$: $p_{1} \to p_{2} \to \hbox{...} \to p_{n}$. Each point $p_{i}$ consists of a geographic location $l$ and a timestamp $t$, $p_{i} = (l,t)$.

\noindent\emph{\textbf{Definition 7: }} \emph{Point of Interest (POI).} A \emph{POI} is a specific point location on the earth, where people may find useful or interesting. Each \emph{POI} contains the name, category, location and other attributes.

\subsection{Framework}
\begin{figure}[ht]
	\centering
	\includegraphics[width=1.0\linewidth]{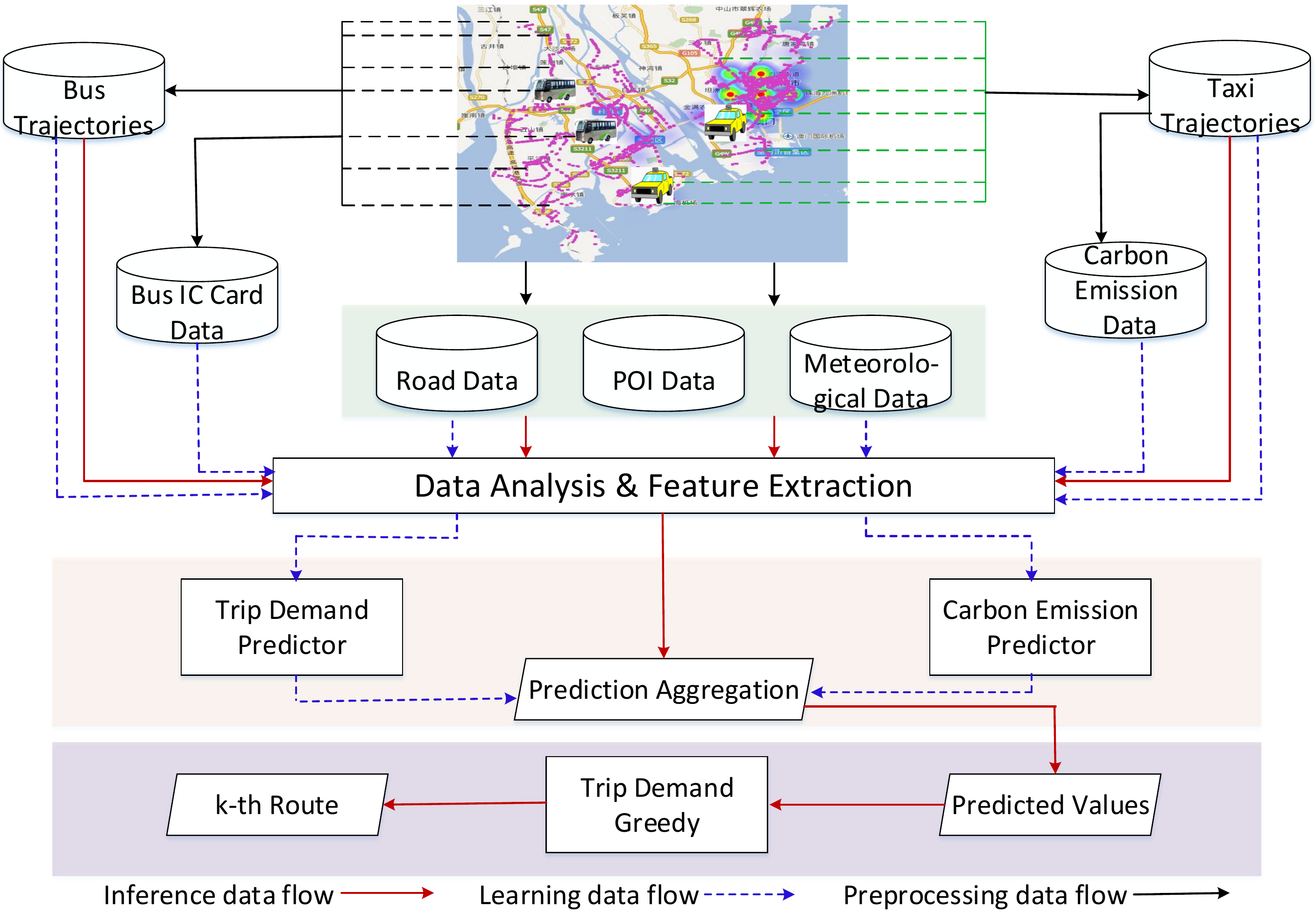}
	\caption{ Framework of our \emph{Clairvoyance} system.}
	\label{pic:Framework}
\end{figure}

Our \emph{Clairvoyance} system provides route recommendations by predicting the future transportation carbon emission and the future bus trip demand according to local conditions, respectively. As shown in Fig. \ref{pic:Framework}, the framework of our system mainly consists of three kinds of data flows, that is preprocessing, learning and inference data flow. 

\noindent \textbf{\emph{Preprocessing data flow:}} In this data flow, we screen out six raw datasets in the city, that is taxi trajectories, bus trajectories, bus \emph{IC} card data, urban road data, \emph{POI}s data and meteorological data. Given the bus \emph{IC} card data, we then estimate the number of bus trips on each bus route during a specified time period and use it to label each bus route. Moreover, based on the taxi trajectories falling in the disjoint grids of the city, we utilize the \emph{Top-Down} approach to calculate the amount of transportation carbon emission in each grid and regard it as a label for each grid.

\noindent \textbf{\emph{Learning data flow:}} In this data flow, we first identify effective features in each grid and each affecting region. The features related to transportation carbon emission are extracted from the taxi trajectories, urban road data, \emph{POI}s data and meteorological data in each grid. Moreover, the features related to bus trip demand are extracted from the bus trajectories, taxi trajectories, \emph{POI}s and meteorological dataset in each affecting region. Then the features of each bus route are generated by incorporating the extracted features in each affecting regions on the route. See section IV for details.

Furthermore, the feature datasets and the labels of each grid and bus route are fed into a \emph{Trip Demand Predictor} and a \emph{Carbon Emission Predictor}, respectively. The \emph{Trip Demand Predictor} is a deep neural network   to predict the future number of people's bus trips on each bus route. The \emph{Carbon Emission Predictor} a three-layer perceptron neural network  to forecast the future transportation carbon emission in each grid. Detailed in section V.

\noindent \textbf{\emph{Inference data flow:}} In this data flow, we respectively extract features from each grid and affecting region. Subsequently, for each bus route, we feed the corresponding features into the trained \emph{Trip Demand Predictor} to forecast the future number of bus trips. For each gird, we feed the extracted features into the trained \emph{Carbon Emission Predictor} to predict the future amount of transportation carbon emission. Based on the two predicted scores, we utilize a greedy mechanism to recommend suitable bus routes for each electric bus that will leave. Detailed in section V.

\noindent\textbf{Problem Formulation}

\noindent Given a collection of bus routes $R$ = $\{r_{1},r_{2},\hbox{...},r_{n} \}$, where each bus route $r$ crosses a collection of grids $G$ = $\{g_{1},g_{2},\hbox{...},g_{n} \}$, a bus trajectory dataset $T_{b}$ and a taxi trajectory dataset $T_t$ passing $R$ and $G$, the \emph{POI}s located in $G$, and a  meteorological dataset, we intend to recommend the suitable bus route $r$ for each electric bus that will depart, based on the inference of $r.BTD$ and $g.W$. The $r.BTD$ is the future number of bus trips on the bus route $r$ and $g.W$ is the future amount of transportation carbon emission in the grid $g$.

\section{Feature Extraction}
In this section, we define and discuss the features that are mined from the heterogeneous data sources in the city. The features related to bus trip demand are classified into three broad categories: \emph{mobility features} which extract information about vehicle movements on road, \emph{geographic features} which integrate information about the \emph{POI} categories and spatial interaction between people and places, and \emph{meteorological features} which exploit knowledge about the weather. Moreover, we identify five kinds of feature sets for the \emph{Carbon Emission Predictor} \cite{lu2017predicting}.

\subsection{Mobility Features: $F_{vm}$}
The daily trip demand of people drives the movement of vehicles on the road. The movement information of vehicles contributes to the analysis of trip demand significantly. However, the running state of vehicles on each road varies dramatically during the same time period. Even on the same road, the vehicle movement state is extremely diverse in different sections of the road. Accordingly, we divide the road into many small affecting regions and extract the unique  movement information in each region. For each affecting region, we identify three types of features: \emph{traffic volume}, \emph{average passing speed} and \emph{standard deviation of speed}. Then we generate the mobility features of the road by aggregating all the identified features in each affecting region on the road.
In the experiments, these features are calculated from the spatial trajectories of taxicabs in Zhuhai.

\noindent \emph{\textbf{Traffic volume: $N(\gamma, T)$.}} The  $N(\gamma, T)$ is the number of vehicles passing through the affecting region $\gamma$ during a specified period of time $T$. Given all the spatial trajectories $\tau_{all}$ generated by vehicles during the period of time $T$, we retrieve all the spatial points on each trajectory $\tau$. If the spatial point $p$ of one trajectory $\tau$ are observed in the affecting region $\gamma$ ($p.l\in\gamma$, $p.t \in T$), we believe that there is one vehicle passing through the affecting region $\gamma$. Even if the same vehicle may pass through the affecting region $\gamma$ many times during the period of time $T$, we only record once. $N(\gamma, T)$ can be formulated as :
\begin{equation}
\label{eq:NumVPT}
N(\gamma, T) = \left| \{\tau \in \tau_{all}: p \in \tau, p.l\in\gamma, p.t \in T \} \right|
\end{equation}
where $p.l$ and $p.t$ represents the geographic location and the timestamp of $p$ respectively.

\noindent \emph{\textbf{Average passing speed: $E(\gamma, T)$.}} The $E(\gamma, T)$ is the average speed of vehicles passing through the affecting region $\gamma$ during the period of time $T$. From all vehicle trajectories $\tau_{all}$ generated over the period of time $T$, we extract the sequences of points that fall in the affecting region $\gamma$.  In the Equation \ref{eq:lenSeq}, $S_\eta$ is the length of one sequence $\eta$, $\eta \in \tau_{all}$. Accordingly, $E(\gamma, T)$ can be obtained by the Equation \ref{eq:Ev}. 
\begin{equation}
\label{eq:lenSeq}
S_\eta = \sum_{p\in\eta } dist(p_{i},p_{i+1}), \qquad p.t \in T
\end{equation}
\begin{equation}
\label{eq:Ev}
E(\gamma, T) =\frac{\sum_{\eta \in \tau_{all}} S_{\eta}}{\sum_{i \in \mathbb{Z}^{+}} \left|p_{i+1}.t - p_{i}.t\right|} , \qquad p.l\in\gamma, p.t \in T
\end{equation}
where the $dist(p_{i},p_{i+1})$ is the traveling distance between any two adjacent points on each spatial trajectory $\tau$.

\noindent \emph{\textbf{Standard deviation of speed: $D(\gamma, T)$.}} The $D(\gamma, T)$ denotes the discrete degree of the passing speed in the affecting region $\gamma$ during the specified period of time $T$. 
\begin{equation}
\begin{aligned}
\label{eq:Dv}
D(\gamma, T) = \sqrt{\frac{\sum_{i \in \mathbb{Z}^{+}} \left[ E(\gamma, T)-p_{i}.v \right] ^2 \cdot \left|p_{i+1}.t - p_{i}.t\right| }{\sum_{i \in \mathbb{Z}^{+}} \left|p_{i+1}.t - p_{i}.t\right|}}
\end{aligned}
\end{equation}
where $p$ is the point on the trajectories that falls in the affecting region $\gamma$ during the period time $T$. $p.v$ denotes the speed of vehicle at the location $p.l$.

\subsection{Geographic Features: $F_{g}$}
The geographic features describe the geographical distributions of \emph{POI}s and the spatial interactions between people and places. \emph{POI} distributions in a region indicate the regional economic and physical attributes, which have high correlation to vehicle trips in the region. Moreover, the spatial interaction states the movement of people between the places. For instance, the high trip needs are more often associated with the residential and industrial areas. Consequently, we identify several correlated features for each affecting region, that is  \emph{entering flow}, \emph{leaving flow} and $POI feature$.

\noindent \emph{\textbf{Entering flow: $f_{in}(\gamma, T)$.}}
The  $f_{in}(\gamma, T)$ is a feature to measure the flow of people into the affecting region $\gamma$ during the period of time $T$. Formally, we utilize a tuple $(O,D)$ to represent the starting point and the end point of a trip. $I$ indicates the total set of the tuple $(O,D)$ of all trips, that are generated in the specific time $T$. 
\begin{equation}
\begin{aligned}
\label{eq:FlowIn}
f_{in}(\gamma, T) = \left| \{(O,D) \in I: O.l \notin \gamma \wedge  D.l\in \gamma \wedge O.t, D.t \in T \} \right|
\end{aligned}
\end{equation}
where $O.l$ and $D.l$ represent their geographic  locations. $O.t$ and $D.t$ denote their timestamps. 

\noindent \emph{\textbf{Leaving flow: $f_{out}(\gamma, T)$.}} 
The  $f_{out}(\gamma, T)$ is a feature to calculate the flow of people out of the affecting region $\gamma$ in the specified time $T$. Similar to Equation \ref{eq:FlowIn}, the leaving flow $f_{out}(\gamma, T)$ out of area $\gamma$ is formulated with Equation \ref{eq:FlowOut}.
\begin{equation}
\begin{aligned}
\label{eq:FlowOut}
f_{out}(\gamma, T) = \left| \{(O,D) \in I: O.l \in \gamma \wedge  D.l\notin \gamma \wedge O.t, D.t \in T \} \right|
\end{aligned}
\end{equation}

\begin{figure}[ht]
	\centering
	\includegraphics[width=0.98\linewidth]{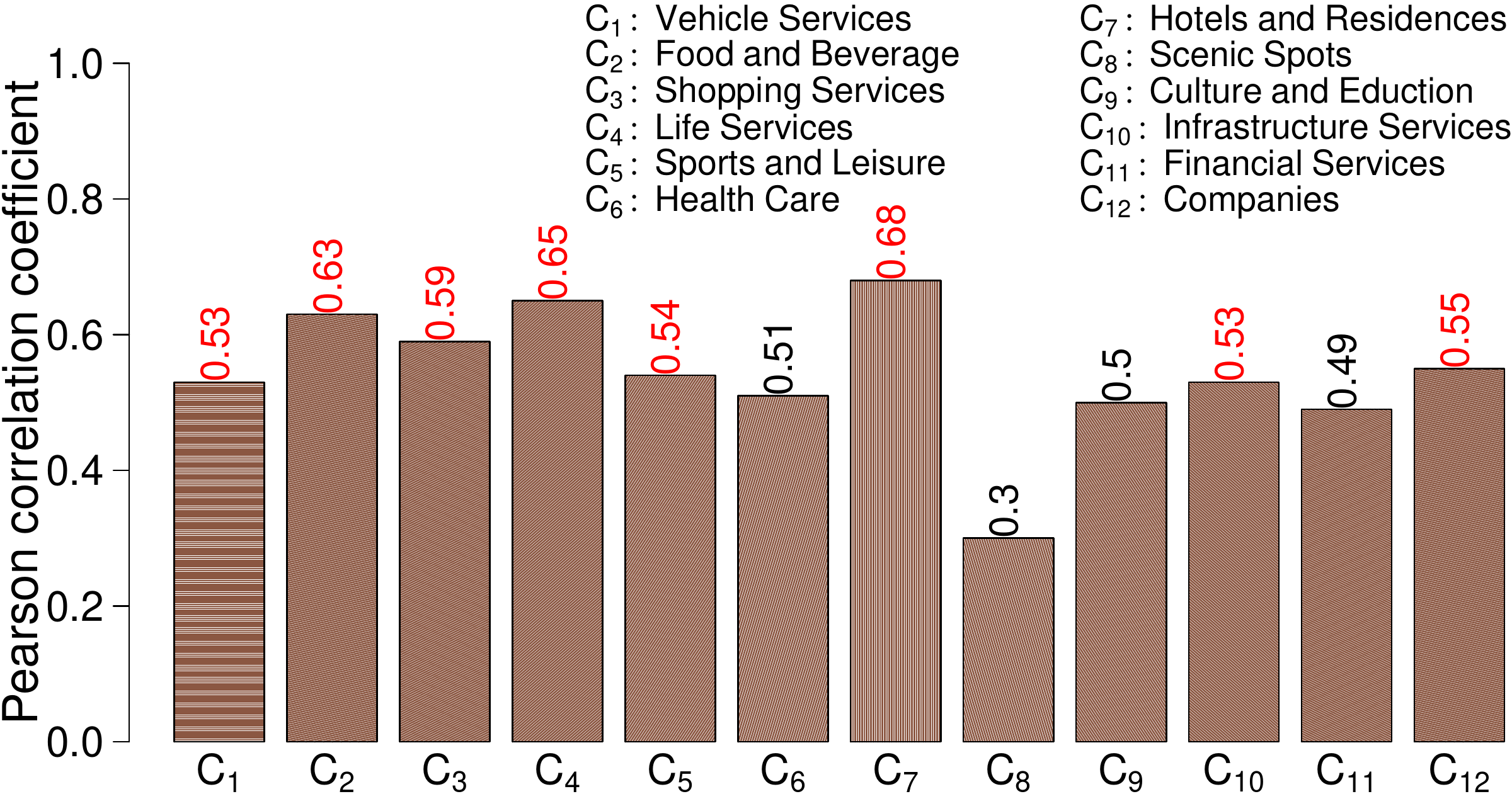}
	\caption{$POI$ categories relevant to vehicle trip demand with the Pearson Correlation Coefficient ($PCC$).}
	\label{pic:POIPCC}
\end{figure}
\noindent \emph{\textbf{POI feature: $P(C, \gamma)$.}}
The \emph{POI} feature $P(C, \gamma)$ is a set of $\theta(C_i,\gamma)$. Each $\theta(C_i,\gamma)$ denotes the number of \emph{POI}s with category $C_i$ in the region $\gamma$. $P(C, \gamma)$ and $\theta(C_i,\gamma)$ are defined as: 
\begin{equation}
\begin{aligned}
\label{eq:POIFeaSet}
P(C,\gamma) = \{ \theta(C_i,\gamma) : C_i \in C \}
\end{aligned}
\end{equation}
\begin{equation}
\begin{aligned}
\label{eq:NumPOI}
\theta(C_i,\gamma) = \left|\{ \varepsilon \in P_i: \varepsilon.l \in \gamma \}\right|
\end{aligned}
\end{equation}
where $P_i$ denotes the set of \emph{POI}s with category $C_i$. $\varepsilon.l$ is the location of the \emph{POI} $\varepsilon$. The set $C$ consists of all the \emph{POI} categories, which are the most relevant to vehicle trip demand. For each category, we use the Pearson Correlation Coefficient ($PCC$) to calculate its correlation with the vehicle trip demand, as displays in Fig. \ref{pic:POIPCC}. For category $C_i$, its $PCC$ value is calculated as:
\begin{equation}
\begin{aligned}
\label{eq:PCC}
\rho(\theta(C_i,\gamma), \nu) = \frac{cov(\theta(C_i,\gamma), \nu)}{\sigma_{\theta(C_i,\gamma)} \sigma_{\nu}}
\end{aligned}
\end{equation}
where $\nu$ is the vehicle trip demand generated in the region $\gamma$ during a specified period of time. $cov$ is the variance of $\theta(C_i,\gamma)$ and $\nu$. $\sigma_{\theta(C_i,\gamma)}$ is the standard deviation of $\theta(C_i,\gamma)$. $\sigma_{\nu}$ is the standard deviation of $\nu$. In the experiments, we select the top-8 \emph{POI} categories that are most relevant to the vehicle trip demand in Zhuhai, $C=\{C_{7},C_{4},C_{2},C_{3},C_{12},C_{10},C_{5},C_{1}\}$.

\begin{table}[!ht]
	\renewcommand{\arraystretch}{1.3}
	\caption{\qquad Typical weather conditions as defined by the Zhuhai \newline Meteorological Bureau in Zhuhai, China. }
	\label{table:WeatherClass}
	\centering
	\begin{tabular}{l l}
		\hline
		\multicolumn{2}{c} { \textbf{ Weather condition categories}} \\
		\hline
		$1:$ Sunny & $11:$Light rain to moderate rain\\
		$2:$ Cloudy & $12:$Moderate rain to heavy rain \\
		$3:$ Shower & $13:$Heavy rain to torrential rain \\
		$4:$ Thunder shower  & $14:$Hot  \\
		$5:$ Light rain  & $15:$Rain to sunny \\ 
		$6:$ Moderate rain   & $16:$Sunny to rain \\ 
		$7:$ Heavy rain   & $17:$Cloudy to sunny \\ 
		$8:$ Torrential rain   & $18:$Cloudy to rain \\ 
		$9:$ Foggy   & $19:$Sunny to cloudy \\ 
		$10:$ Haze   & $20:$Rain to cloudy \\ 
		\hline
	\end{tabular}
\end{table}

\subsection{Meteorological Features: $F_{m}$}
Changes in meteorological conditions could have several potential consequences for people's travel demand on a global and regional scale. Especially the weather changes have a significant impact on the traffic pattern, the purpose of the trip and the travel time \cite{koetse2009weatherimpact}. For instance, the increased frequency of extreme precipitation events may substantially reduce the number of trips, particularly those for leisure purpose. Accordingly, we give special attention to the impact of weather on people's trip in our study. We identify three features, that is \emph{highest temperature}, \emph{lowest temperature}, and \emph{weather condition}. More specifically, the $F_{m}$ of each bus route can be formulated as follows \ref{eq:F_m}:
\begin{equation}
\begin{aligned}
\label{eq:F_m}
F_{m}(r) = \{ (f_{h},f_{o},f_{w}) : f_{w} \in M  \}
\end{aligned}
\end{equation}
where $F_{m}(r)$ represents the meteorological feature set of the bus route $r$. $M$ is the set of weather condition categories. The weather conditions are classified into 20 categories, as displayed in the table \ref{table:WeatherClass}. $f_{h}$ is the \emph{highest temperature}. $f_{o}$ is the \emph{lowest temperature}. $f_{w}$ is the \emph{weather condition}.

\begin{figure}[ht]
	\centering
	\includegraphics[width=1.0\linewidth]{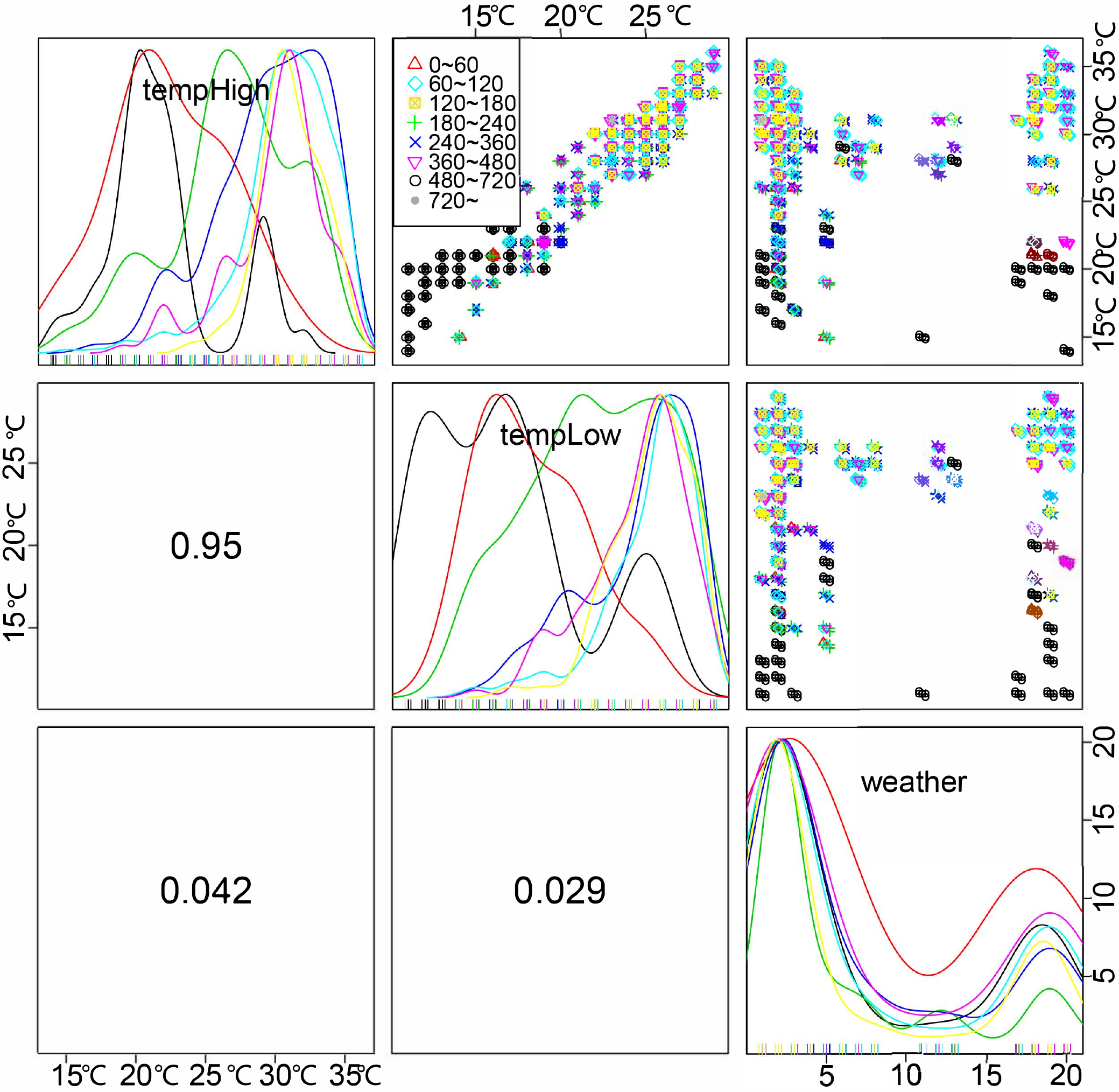}
	\caption{ Correlation matrix between meteorological features and bus trip demand.}
	\label{pic:busMeteoFea}
\end{figure}

Fig. \ref{pic:busMeteoFea} displays the correlation matrix between the meteorological features and the bus trip demand, based on the meteorological data in Zhuhai from January to October, 2015. In this figure, each row and each column represent one kind of meteorological features. The \emph{tempHigh}, \emph{tempLow} and \emph{weather} denote $f_h$, $f_o$ and $f_w$ respectively. As depicted in the legends, the descriptors and colors indicate the trip number ranges. We have eight trip number ranges, in which `0$\sim$60' is a range of 0 to 60 trips, and `720$\sim$' represents more than 720 trips. The plots in upper triangular portion indicate the variation of people's bus trips under different meteorological features.  Apparently, the bigger trip ranges often occur when the \emph{tempHigh} and \emph{tempLow} are both relatively low on the sunny or cloudy day. For instance, the plot on the first row and the second column suggests the `480$\sim$720' always occur when the \emph{tempLow} is less than 20 $^{\circ}$C  and \emph{tempHigh} is less than 25 $^{\circ}$C. The density curves of the trip ranges are manifested in the diagonal portion of the correlation matrix. As shown in the plot on the third row and the third column, the density curves vary noticeably with the weather conditions. Moreover, the plots in lower triangular portion  state clearly the Pearson Correlation Coefficient between the meteorological features. As an illustration, the value 0.95 in the plot on the second row and the first column indicates the Pearson Correlation Coefficient between the$f_h$ and $f_o$.

\section{Learning and Inference}
Based on the features extracted from urban big data in the previous part, our \emph{Clairvoyance} system is trained to predict the number of people's bus trips on each bus route and forecast the transportation carbon emission in each grid during a future period of time. Given the predicted information, we propose a greedy strategy to recommend  suitable bus routes for electric buses that will depart.

\subsection{Two-task Forecasting}
Two main parts of our \emph{Clairvoyance} system are \emph{Trip Demand Predictor} and \emph{Carbon Emission Predictor}. 
As presented in Algorithm  \ref{al:Co-forecastingModel}, 
the \emph{Trip Demand Predictor} forecasts the future number of people's bus trips (\emph{BTD}) on each bus route, utilizing a deep neural network pre-trained by stacking four layers of denoising autoencoders (\emph{SDAE-4}). The \emph{Carbon Emission Predictor} predicts the future transportation carbon emission (\emph{TCE}) in each grid, using a three-layer perceptron neural network (\emph{PNN-3}).
Moreover, we train them based on the datasets of $(F(r),\mu_r)$ and $(F(g),\nu_g )$, respectively. 
$F(r)$ is a feature set of $F_{vm}$, $F_{g}$, and $F_{m}$ on the bus route $r$, and $\mu_r$ is the corresponding \emph{BTD} value. $F(g)$ and $\nu_g$ are another feature set and the corresponding \emph{TCE} value, which are described in detail in [28]. 
Subsequently, given the two trained predictors, we forecast separately the \emph{BTD} values on each bus route and the \emph{TCE} values in each grid throughout the city within a future period of time.

\begin{algorithm}[ht]
	\caption{ Two-task Forecasting }
	\label{al:Co-forecastingModel}
	\renewcommand{\algorithmicrequire}{\textbf{Input:}}
	\renewcommand{\algorithmicensure}{\textbf{Output:}}
	
	\begin{algorithmic}[1] 
		\Require A training dataset $D_{\emph{SDAE}}$ of ($F(r)$,$\mu_r$), where $F(r)$ is a feature set of $F_{vm}$, $F_{g}$, and $F_{m}$ and $\mu_r$ is the \emph{BTD} value on the route $r$; A training dataset $D_{\emph{PNN}}$ of ($F(g)$,$\nu_g$), where the feature set $F(g)$ and the \emph{TCE} value $\nu_g$ are extracted in the grid $g$; A specified period of time interval $T$.
		\Ensure The set $\mu_{R}$ of \emph{BTD} values on all bus routes; the set $\nu_{G}$ of \emph{TCE} values in all grids.

		\State \emph{SDAE}-$4$ $\gets$ $build$ $a$ $4$-$hidden$-$layer$ $neural$ $network$ $pre$-$trained$ $by$ $stacking$ $denoising$ $autoencoders$. 
		\State \emph{PNN}-$3$ $\gets$ $design$ $a$ $3$ $layer$ $perceptron$ 

		\For{ $small$ $batch$ $in$ $D_{\emph{SDAE}}$}
		\State$\emph{MiB}_{\emph{SDAE}}$ $\gets$ $the$ $small$ $batch$ $in$ $D_{\emph{SDAE}}$
		\State\emph{SDAE}-$4$ $\gets$ \emph{SDAE}-$4$$.Learning ({\emph{MiB}_\emph{SDAE}})$
		\EndFor

		\For{ $small$ $batch$ $in$ $D_{\emph{PNN}}$}
		\State$\emph{MiB}_{\emph{PNN}}$ $\gets$ $the$ $small$ $batch$ $in$ $D_{\emph{PNN}}$
		\State$PNN$-$3$ $\gets$ $PNN$-$3$$.Learning (\emph{MiB}_{\emph{PNN}})$
		\EndFor
		\State $\mu_{R}$ $\gets$ $using$ $the$ $trained$ \emph{SDAE}-$4$ $to$ $predict$ \emph{BTD} $values$ $\{\mu_{r_{1}},\mu_{r_{2}},\hbox{...},\mu_{r_{n}} \}$ $on$ $all$ $bus$ $routes$ $R$ $in$ $the$ $future$ $period$ $of$ $time$ $T$, $R$ = $\{r_{1},r_{2},\hbox{...},r_{n} \}$. 
		\State $\nu_{G}$ $\gets$ $using$ $the$ $trained$ \emph{PNN}-$3$ $to$ $predict$ \emph{TCE} $values$ $\{\nu_{g_{1}},\nu_{g_{2}},\hbox{...},\nu_{g_{n}} \}$ $in$ $all$ $grids$ $G$ $in$ $the$ $future$ $period$ $of$ $time$ $T$, $G$ = $\{g_{1},g_{2},\hbox{...},g_{n} \}$. 

\noindent \Return  $\mu_{R}$ and $\nu_{G}$
	\end{algorithmic}
\end{algorithm}

\subsection{Trip Demand Predictor}
\begin{figure}[ht]
	\centering
	\includegraphics[width=1.0\linewidth]{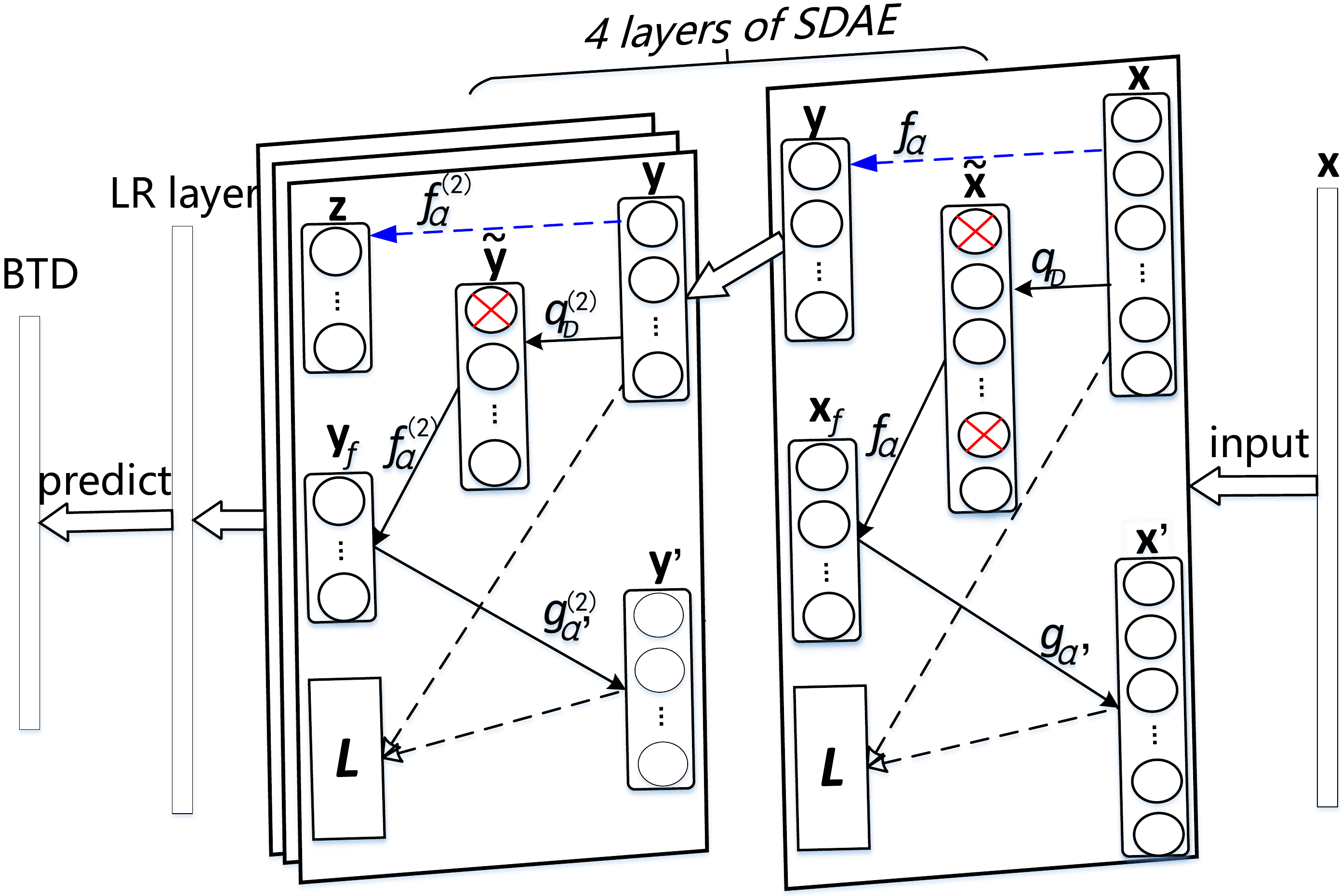}
	\caption{ The deep neural network pre-trained by stacking four layers of denoising autoencoders (\emph{SDAE-4}). In each \emph{SDAE} layer \cite{vincent2010stacked}, the black solid and broken arrows represent the flow of learning process (black solid first), and the blue broken arrows represent the flow of inference process.}
	\label{pic:SDAEstructure}
\end{figure}

As depicted in Fig. \ref{pic:SDAEstructure}, the \emph{Trip Demand Predictor} is composed of four main parts: input, four layers of denoising autoencoders(\emph{SDAE}), \emph{logistic regression} (\emph{LR}) layer, and output. The initial input vector $\mathbf{x}$ consists of $F_{vm}$, $F_{g}$, $F_{m}$ and the corresponding \emph{BTD} label. The four layers of \emph{SDAE} build a deep neural network architecture by feeding the latent representation found on the layer below as input to the current layer. In each layer of \emph{SDAE}, the black solid and broken arrows represent the learning process, and the blue broken arrows represent the inference process. In the \emph{LR} layer, a multi-class logistic regression method is used to generate the predicted \emph{BTD} values. The input of \emph{LR} layer is the output of the last level \emph{SDAE}.

The training of our deep neural network \emph{SDAE-4} includes two stages, that is \emph{unsupervised pre-training} and \emph{supervised fine-tuning}. The \emph{unsupervised pre-training} of the deep network is done by the four layers of \emph{SDAE}. Each denoising autoencoder is trained to minimize the input reconstruction error. As shown in Fig. \ref{pic:SDAEstructure}, the first level \emph{SDAE} takes an input vector $\mathbf{x}$. Subsequently, the initial input $\mathbf{x}$ is partially destroyed to get a corrupted input $\mathbf{\tilde{x}}$, using the means of a stochastic mapping $\mathbf{\tilde{x}} \sim q_D(\mathbf{\tilde{x}}|\mathbf{x})$. In the experiments, in order to obtain the corrupted input, we randomly assign a fixed number of elements in each input vector to zero. Then, the corrupted input $\mathbf{\tilde{x}}$ is mapped to a hidden representation $\mathbf{x}_{f}$, using the sigmoid non-linearity:
\begin{equation}
\begin{aligned}
\label{eq:Encoder}
\mathbf{x}_{f} &= f_{\alpha}(\mathbf{\tilde{x}}) = s(\mathbf{W}\mathbf{\tilde{x}} + \mathbf{b}),
\end{aligned}
\end{equation}
where $\alpha$ denotes the parameter set of $f_{\alpha}$, $\alpha=\{ \mathbf{W}, \mathbf{b}\}$. $\mathbf{W}$ is a $d'\times d$ weight matrix. $\mathbf{b}$ is an offset vector of dimensionality $d'$. $s$ is the sigmoid function, which can be formulated as follows:
\begin{equation}
\begin{aligned}
\label{eq:sigmoidx}
&s(x) = \frac{1}{1+e^{-x}} \\
&s(\mathbf{x}) = (s(x_1),s(x_2),\cdots, s(x_d))^{\rm T}.
\end{aligned}
\end{equation}
where $s(x_1)$,$s(x_2)$, and $s(x_d)$ are the elements of the input vector $\mathbf{x}$.
When the hidden representation $\mathbf{x}_{f}$ is learnt, we reconstruct the initial input $\mathbf{x}$ by the following Equation \ref{eq:Decoder}, to generate produce reconstruction $\mathbf{x}'$.
\begin{equation}
\begin{aligned}
\label{eq:Decoder}
\mathbf{x}' &= g_{\alpha'}(\mathbf{x}_{f}) = s(\mathbf{W}'\mathbf{x}_{f} + \mathbf{b}'),
\end{aligned}
\end{equation}
where $\alpha'$ denotes the parameter set of $g_{\alpha'}$, $\alpha' = \{ \mathbf{W}', \mathbf{b}'\}$. The weight matrix $\mathbf{W}'$ is the transpose of the matrix $\mathbf{W}$, $\mathbf{W}'= \mathbf{W}^{\rm T}$. Accordingly, for each training $\mathbf{x}^{(i)}$, we can get the corresponding $\mathbf{x}_{f}^{(i)}$ and reconstruction $\mathbf{x}'^{(i)}$. Meanwhile, in order to minimize the \emph{average reconstruction error}, the parameters of denoising autoencoder are optimized by the following Equation \ref{eq:miniError}.
\begin{equation}
\begin{aligned}
\label{eq:miniError}
\alpha^\star,\alpha'^{\star} &= \arg \min_{\alpha,\alpha'} \frac{1}{n} \sum_{i=1}^{n} (\mathbf{x}^{(i)}-\mathbf{x}'^{(i)})^2  \\
&= \arg \min_{\alpha,\alpha'} \frac{1}{n} \sum_{i=1}^{n} (\mathbf{x}^{(i)}-g_{\alpha'}(f_{\alpha}(\mathbf{\tilde{x}}^{(i)})))^2,
\end{aligned}
\end{equation}
where $n$ is the number of input samples. In the experiments, we measure the reconstruction errors by \emph{mean squared error}.  The \emph{mean squared error} can be formulated as follows:
\begin{equation}
\begin{aligned}
\label{eq:lossFunc}
L(\mathbf{x}, \mathbf{x}') &= \frac{1}{n} \sum_{i=1}^{n}(\mathbf{x}^{(i)}-\mathbf{x}'^{(i)})^2.
\end{aligned}
\end{equation}

After the first-level \emph{SDAE} is trained, its mapping function $f_\alpha$ will be used on clean inputs, as represented by broken blue arrows. Utilizing the learnt function $f_\alpha$, we map the uncorrupted input $\mathbf{x}$ to produce the hidden representation $\mathbf{y}$, which will serve as clean input to the second-level \emph{SDAE}. Subsequently, the remaining layers of \emph{SDAE} are trained, similar to the process in the first-level \emph{SDAE}. 

The \emph{supervised fine-tuning} of \emph{SDAE-4} will be done, after the four layers of \emph{SDAE} are trained. In the $LR$ layer, we generate the probability that the predicted $Y$ is class $c_i$, by the formula:
\begin{equation}
\begin{aligned}
\label{eq:logisticRegression}
P(Y = c_i|\mathbf{x}_L,\mathbf{W}_L,\mathbf{b}_L) &= softmax_{c_i}(\mathbf{W}_L \mathbf{x}_L + \mathbf{b}_L)  \\
&= \frac{e^{W_{L}^{(c_i)}\mathbf{x}_L + b_{L}^{(c_i)}}}{\sum_{j=1}^{k} e^{W_{L}^{(c_j)}\mathbf{x}_L + b_{L}^{(c_j)}}},
\end{aligned}
\end{equation}
where $k$ denotes the number of categories. $\mathbf{x}_L$, $\mathbf{W}_L$, and $\mathbf{b}_L$ are separately the input, weight matrix, and bias vector of the multi-class logistic regression method in the \emph{LR} layer, given the input $\mathbf{x}$ of \emph{SDAE-4}. Subsequently, the output $y_{pred} $ of the $LR$ layer is the class whose probability is maximal, specifically: 
\begin{equation}
\begin{aligned}
\label{eq:LRprediction}
y_{pred} = \arg \max_{i}P(Y = c_i|\mathbf{x}_L,\mathbf{W}_L,\mathbf{b}_L).
\end{aligned}
\end{equation}
Moreover, the loss function of our \emph{SDAE-4} is the \emph{negative log-likelihood} (\emph{NLL}), which is formulated as:
\begin{equation}
\begin{aligned}
\label{eq:likelihood}
NLL(\beta, \mathbf{D}^{(s)}) = -\sum_{i=1}^{|\mathbf{D}^{(s)}|}log(P(Y = \mathbf{y}^{(i)}|\mathbf{x}^{{(i)}},\beta)), 
\end{aligned}
\end{equation}
where $\beta$ is the set of all parameters in the \emph{SDAE-4}. $\mathbf{D}^{(s)}$ denotes the input dataset. $\mathbf{y}^{(i)}$ indicates the corresponding \emph{BTD} label. To optimize our \emph{SDAE-4}, \emph{stochastic gradient descent} is applied to minimize \emph{NLL}.

\subsection{Recommending Bus Route}
Our aim is to recommend $k$ routes, so that the transport of electric buses on these routes will meet people's trips needs and cut transportation carbon emission to the maximum extent.
Given the pre-trained \emph{SDAE-4}, we obtain a set of the \emph{BTD} values $\mu_{R}$ on all bus routes $R$ and a set of the \emph{TCE} values $\nu_{G}$ in all grids $G$ during a period of time $T$ in the future. Meanwhile, we calculate the \emph{TCE} value $\vartheta_{r}$ of the bus route $r$, by adding the \emph{TCE} values in the all grids on the route $r$. 
Subsequently, we normalize the $\mu_{R}$ and $\vartheta_{R}$, as the following:
\begin{equation}
\begin{aligned}
\label{eq:BTD_R_norm}
\mu_{R}' = \{ \mu_{r_i}' | \mu_{r_i}' = \frac{\mu_{r_i}-\mu_{min}}{\mu_{max} - \mu_{min}} , r_{i} \in R  \},
\end{aligned}
\end{equation}
\begin{equation}
\begin{aligned}
\label{eq:TCE_R_norm}
\vartheta_{R}' = \{ \vartheta_{r_i}' | \vartheta_{r_i}' = \frac{\vartheta_{r_i}-\vartheta_{min}}{\vartheta_{max} - \vartheta_{min}} , r_{i} \in R  \},
\end{aligned}
\end{equation}
where $\mu_{max}$ and $\mu_{min}$ are the maximum and minimum values in $\mu_{R}$, $\mu_{R}=\{\mu_{r_1}, \mu_{r_2},\hbox{...},\mu_{r_n}\}$. $\vartheta_{max}$ and $\vartheta_{min}$ are  the maximum and minimum values in $\vartheta_{R}$, $\vartheta_{R}=\{\vartheta_{r_1}, \vartheta_{r_2},\hbox{...},\vartheta_{r_n}\}$. $\mu_{r_i}$ is the \emph{BTD} value on the bus route $r_i$ in a future period of time $T$. $\vartheta_{r_i}$ is the future total \emph{TCE} value of the grids that the bus route $r_i$ crosses. Accordingly, when the value of $\mu_{r}'$ plus $\vartheta_{r}'$ is larger enough, people's bus trips and transportation carbon emissions will peak in the area that the bus route $r$ crosses. Based on the future peak information, we recommend $k$ bus routes for each electric bus that will depart, specifically:
\begin{equation}
\begin{aligned}
\label{eq:optRoute}
O_{r} = \{r_{i}:Max_{top_k}(\mu_{r_{i}}' + \vartheta_{r_{i}}'), \mu_{r_{i}}' \in \mu_{R}', \vartheta_{r_{i}}' \in \vartheta_{R}' \}.
\end{aligned}
\end{equation}
where $Max_{top_k}$ means obtaining the \emph{top-k} maximum value.

\section{Experiments}
In this section, we will introduce the source datasets used in the experiments. Several advanced and baseline methods are selected to compare with our proposed methods. Subsequently, we present experimental results showing the performance of our proposed methods and baseline methods.
\subsection{Datasets}
In the experiments, we use five real world datasets collected in Zhuhai, China. The statistics of datasets are displayed in Table \ref{table:datasets}. 

\noindent $\mathbf{1}$) \textbf{\emph{Taxi trajectories:}} 
The dataset was generated by over $3,000$ taxicabs in Zhuhai from July 1st to October 14th, 2015. Specifically, each \emph{GPS} record in the taxi trajectory dataset is produced every 10 seconds and consists of seven parts, that is \emph{taxi ID}, \emph{record time}, \emph{latitude}, \emph{longitude}, \emph{speed}, \emph{direction} and \emph{passenger status}. Furthermore, in one day all the taxicabs can drive a mileage of about 0.116 million kilometers and generate over 38 million records. More than 4.5 million vehicle trips will occur one day, and the trips of taxicabs and small passenger cars account for about 31.8\%, in accordance with the annual report of the 2014 traffic development in Zhuhai. Accordingly, this source dataset is large enough to reflect local traffic patterns there \cite{zheng2013u_air}.

\noindent $\mathbf{2}$) \textbf{\emph{Bus trajectories:}} 
The dataset was produced by about $1,540$ buses in Zhuhai from July 1st to October 14th, 2015. More specifically, each record in the bus trajectory dataset is composed of eight main components, that is \emph{bus ID}, \emph{record time}, \emph{latitude}, \emph{longitude}, \emph{speed}, \emph{last position}, \emph{current bus stop}, and \emph{route name}. When each bus passes any bus stop on the routes, two records will be generated. Moreover, there are about 168 bus routes in Zhuhai. 

\noindent $\mathbf{3}$) \textbf{\emph{IC card data:}} 
The dataset was produced by \emph{IC} cardholders on all buses in Zhuhai from January 1st to October 31st, 2015. Each record in the \emph{IC} card dataset consists of four main parts, that is \emph{bus route ID}, \emph{bus ID}, \emph{consumption date}, and \emph{card ID}. Moreover, there are more than 600 thousand user consumption records generated in one day in Zhuhai.

\begin{figure}[ht]
	\centering
	\includegraphics[width=1.0\linewidth]{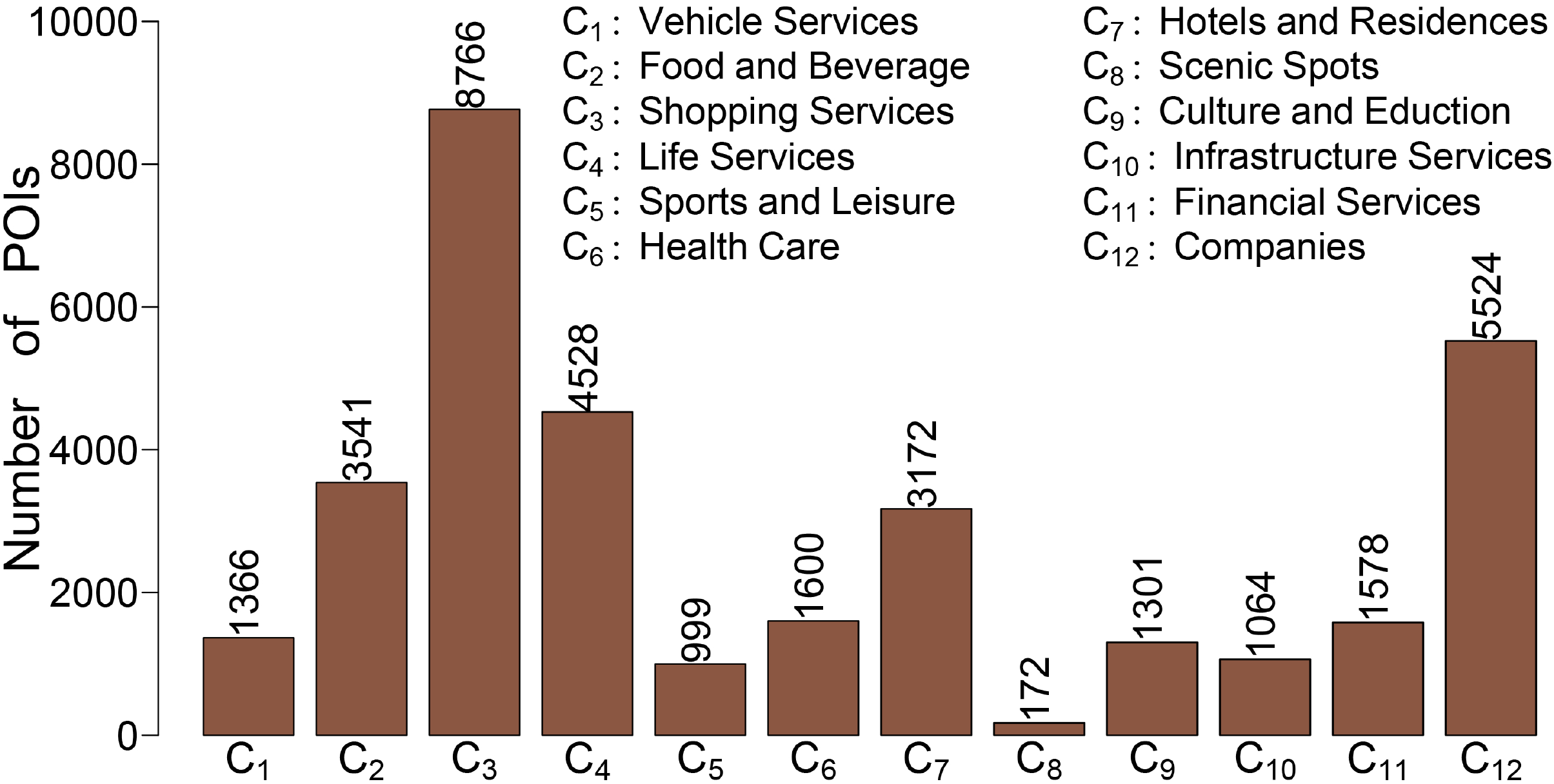}
	\caption{ \emph{POI} categories with the number of records in Zhuhai, China.}
	\label{pic:numPOI}
\end{figure}

\noindent $\mathbf{4}$) \textbf{\emph{POIs:}} 
The dataset was extracted from the \emph{Auto Navi MAP} in Zhuhai. Specifically, each record in the \emph{POI} dataset mainly contains five parts, that is \emph{POI ID}, \emph{name}, \emph{type}, \emph{latitude}, and \emph{longitude}. In addition, according to the type of each \emph{POI}, the whole \emph{POI}s can be divided into twelve categories, and each category contains different number of records, as displayed in Fig. \ref{pic:numPOI}. 

\noindent $\mathbf{5}$) \textbf{\emph{Meteorological data:}} The dataset was collected from \emph{National Meteorological Center of China Meteorological Administration} \cite{httpNMC} in Zhuhai from January 1st to October 31st, 2015. Each record in this dataset is composed of three main messages, that is \emph{highest temperature}, \emph{lowest temperature}, and \emph{weather condition}.

\begin{table}[!ht]
	\renewcommand{\arraystretch}{1.3}
	\caption{Statistics of the source datasets in Zhuhai, China.}
	\label{table:datasets}
	\centering
	\begin{tabular}{|l|l|c|}
		\hline
		\multicolumn{2}{|c|}{\textbf{Data sources}} & \textbf{Zhuhai}  \\
		\hline
		Taxi Trajectories & \emph{GPS} Trajectories& 2015/07/01-2015/10/14\\
		\hline
		Bus Trajectories & \emph{GPS} Trajectories& 2015/07/01-2015/10/14\\
		\hline
		\emph{IC} Card Data & \emph{IC} Card Records & 2015/01/01-2015/10/31\\
		\hline
		\emph{POI}s& 2015 & 38815 \\
		\hline
		\multirow{2}{*}{Meteorology}
		& Hours & 8760 \\
		\cline{2-3}
		& Time Spans & 2015/01/01-2015/10/31 \\
		\hline
		\multicolumn{2}{|c|}{ Gird Sizes} & 5 $\times$ 5 $km$  \\
		\hline
		\multicolumn{2}{|c|}{ Affecting Region Sizes} & 0.5 $\times$ 0.5 $km$  \\
		\hline
	\end{tabular}
\end{table}
\subsection{Comparison Methods}
We compare our \emph{SDAE-4} with five baselines, that is logistic regression, support vector machine, restricted Boltzmann machine, deep belief networks, artificial neural network. Meanwhile, we compare our 3-layer perceptron neural network with five other baselines \cite{lu2017predicting}. 

\noindent $\mathbf{1}$) \textbf{\emph{Logistic Regression:}} \emph{LR} is a linear classifier. Moreover, \emph{LR} is often employed in an regression analysis to describe the relationship between the categorical response variable and one or more explanatory variables \cite{harrell2015regression} \cite{LR2013Hosmer} . Mathematically, the \emph{LR} model can be formulated by the Equation \ref{eq:logisticRegression} and \ref{eq:LRprediction}, which are parametrized by a initial weight matrix and a bias vector.

\noindent $\mathbf{2}$) \textbf{\emph{Support Vector Machine:}} \emph{SVM} \cite{chang2011libsvm} is a supervised learning model. In this baseline, we apply an \emph{SVM} with \emph{RBF} kernel (\emph{SVM}$_{rbf}$) for classification and regression analysis. Furthermore, the \emph{radial basis function} (\emph{RBF}) kernel on two sample vectors $\mathbf{x}^{(r)}$ and $\mathbf{x'}^{(r)}$ can be formulated as: 
\begin{equation}
\begin{aligned}
\label{eq:RBF}
K(\mathbf{x}^{(r)},\mathbf{x'}^{(r)}) = \exp(-\frac{\parallel \mathbf{x}^{(r)}-\mathbf{x'}^{(r)}\parallel^2 }{2\sigma^2}), 
\end{aligned}
\end{equation}
where $\parallel \mathbf{x}^{(r)}-\mathbf{x'}^{(r)}\parallel^2$ denotes the the squared Euclidean distance between $\mathbf{x}^{(r)}$ and $\mathbf{x'}^{(r)}$. $\sigma$ is a free parameter.

\noindent $\mathbf{3}$) \textbf{\emph{Restricted Boltzmann Machine:}} \emph{RBM} is a variant of Boltzmann machines \cite{hinton2012practical} \cite{ackley1985learning}. In this baseline, we make use of \emph{RBM} to learn a probability distribution $P$, based on our training dataset, specifically:
\begin{equation}
\begin{aligned}
\label{eq:RBM_E} 
E(v,h) = -a^{\rm T}v - b^{\rm T}h - v^{\rm T}W^{(m)}h,
\end{aligned}
\end{equation}
\begin{equation}
\begin{aligned}
\label{eq:RBM_P} 
P(v,h) = \frac{e^{-E(v,h)}}{\sum e^{-E(v,h)}},
\end{aligned}
\end{equation}
where $v$ and $h$ denote the visible vector and hidden vector, respectively. $a$ and $b$ separately represent the offset vectors of the visible and hidden layers. $a^{\rm T}$ and $b^{\rm T}$ indicate the transpose of $a$ and $b$. $W^{(m)}$ is a matrix of weights connecting hidden and visible units in layer $m$. 

\noindent $\mathbf{4}$) \textbf{\emph{Artificial Neural Network:}}
\emph{ANN} \cite{haykin2009neural} is designed as a one-hidden-layer \emph{Multi-Layer Perceptron (MLP)} in the experiment. Mathematically, the single-hidden-layer \emph{MLP} is formulated as:
\begin{equation}
\begin{aligned}
\label{eq:ANN_MLP} 
f(\mathbf{x}) &= softmax(W^{(2)}a^{(2)} + b^{(2)}) \\
&= softmax(W^{(2)}(s(W^{(1)}\mathbf{x} + b^{(1)})) + b^{(2)}),
\end{aligned}
\end{equation}
where $a^{(2)}$ denotes the output vector of the hidden layer. $W^{(1)}$ and $W^{(2)}$ are separately matrices of weights connecting input and hidden units, and connecting hidden and output units. $b^{(1)}$ and $b^{(2)}$ respectively represent the bias vectors for hidden and output units. $s$ represents the logistic $sigmoid$ function. 

\noindent $\mathbf{5}$) \textbf{\emph{Deep Belief Networks:}}
\emph{DBN} is a deep network model pre-trained by stacking \emph{RBM}s \cite{sarikaya2014application} \cite{lee2009convolutional}. Furthermore, \emph{DBN} trys to learn the joint probability distribution between the $\ell$ hidden vector $h^k$ and observed vector $\mathbf{x}$, specifically:
\begin{equation}
\begin{aligned}
\label{eq:DBN} 
P(\mathbf{x},h^1,\cdots,h^{\ell}) = (\prod_{k=0}^{\ell-2}P(h^{k}\mid h^{k+1})) P(h^{\ell-1}, h^{\ell}),
\end{aligned}
\end{equation}
where $P(h^{k}\mid h^{k+1})$ denotes a conditional distribution for the visible units given the hidden units of the \emph{RBM} at level $k+1$. $ P(h^{\ell-1}, h^{\ell})$ represents the visible-hidden joint distribution in the top-level \emph{RBM}.
\subsection{Results}
\noindent $\mathbf{1}$)\quad\textbf{\emph{Evaluating with the fusion of different features:}}

\begin{figure}[ht]
	\subfigure[Predicting the \emph{BTD} on the bus routes in the next 1 hour. ]{
		\label{pic:accuracyH1}
		\includegraphics[width=1.6in]{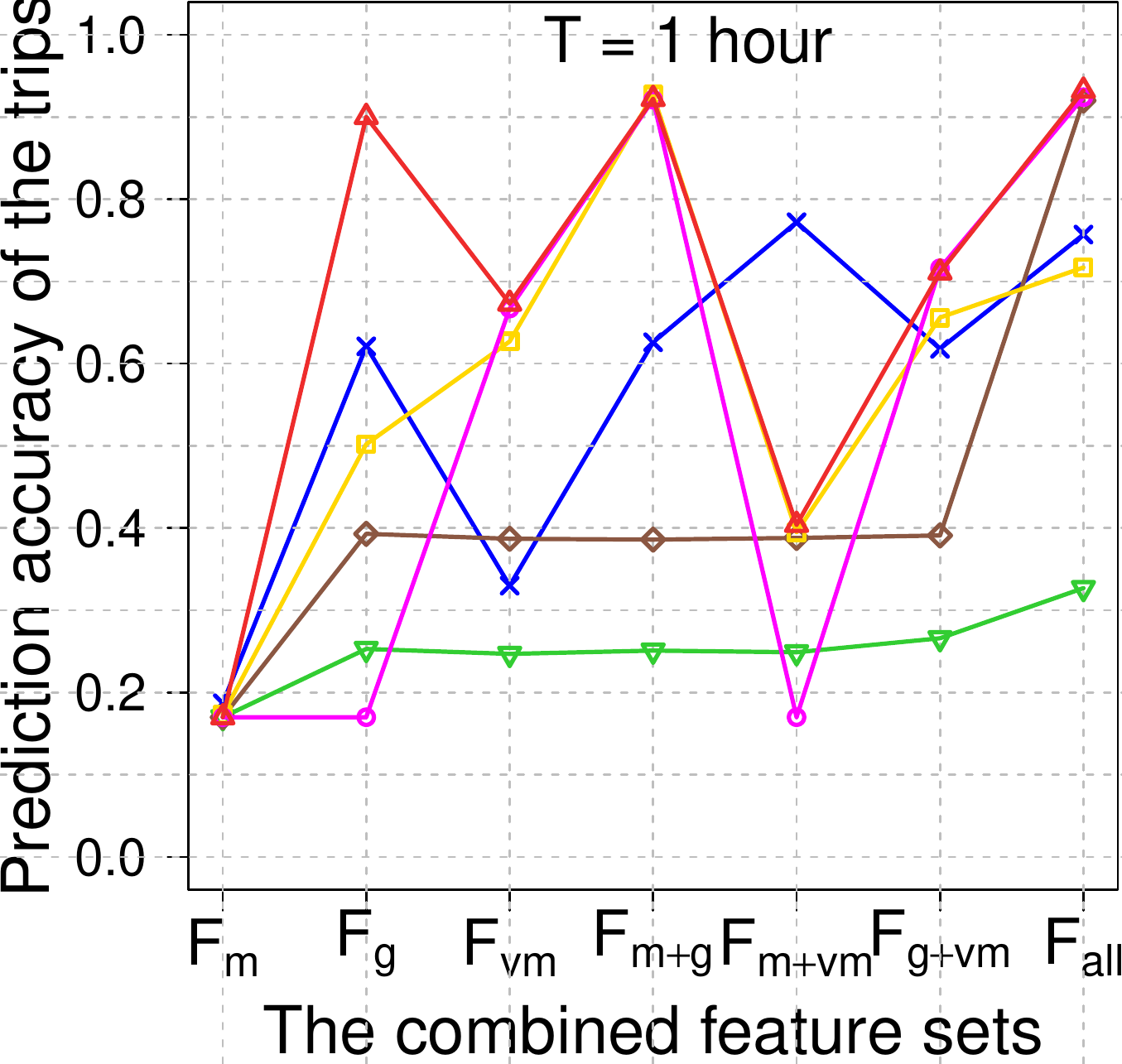}}
	\hspace{0in}
	\subfigure[Predicting the \emph{BTD} on the bus routes in the next 2 hours.]{
		\label{pic:accuracyH2} 
		\includegraphics[width=1.6in]{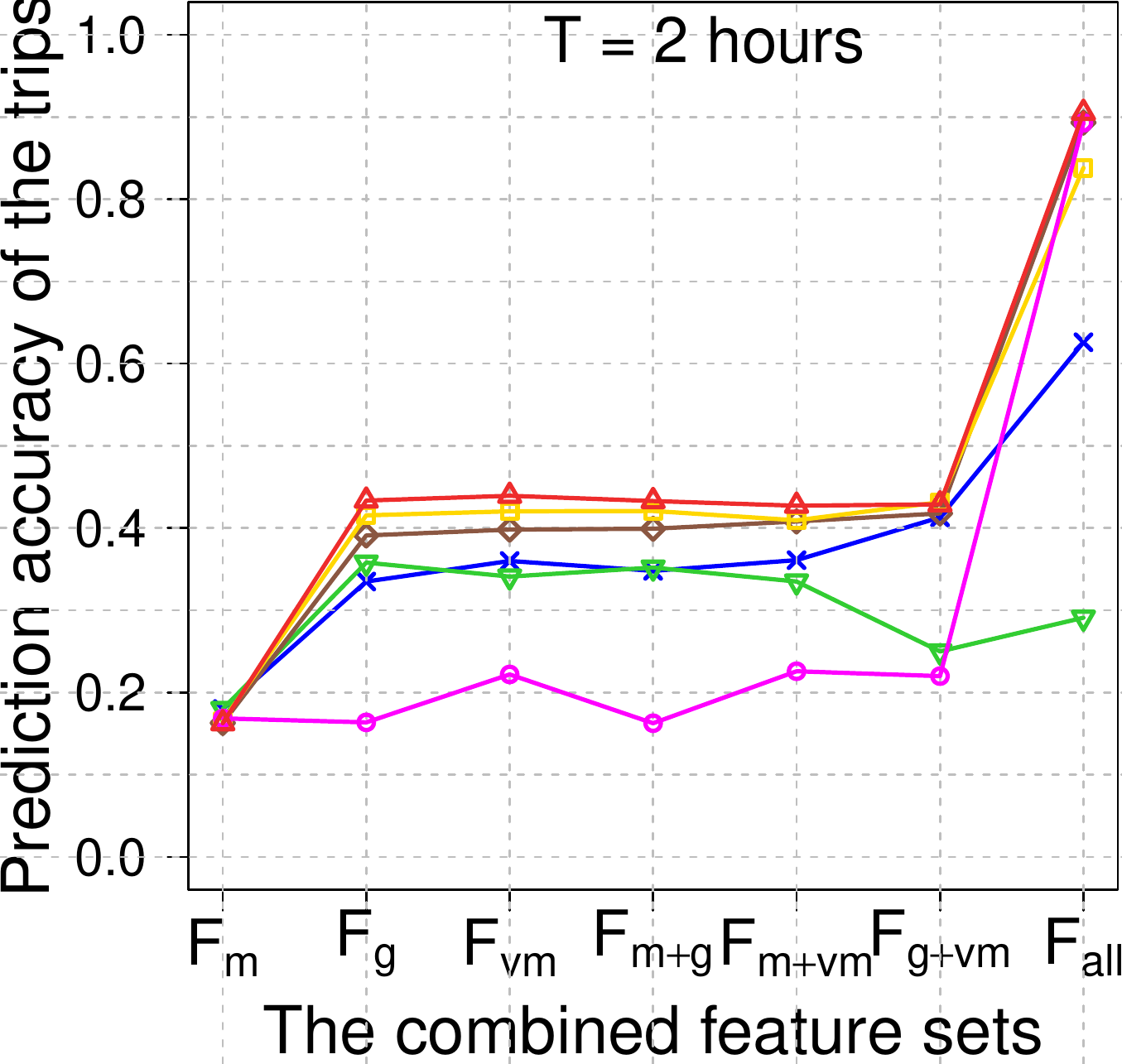}}
	\subfigure[Predicting the \emph{BTD} on the bus routes in the next 3 hours.]{
		\label{pic:accuracyH3}
		\includegraphics[width=1.6in]{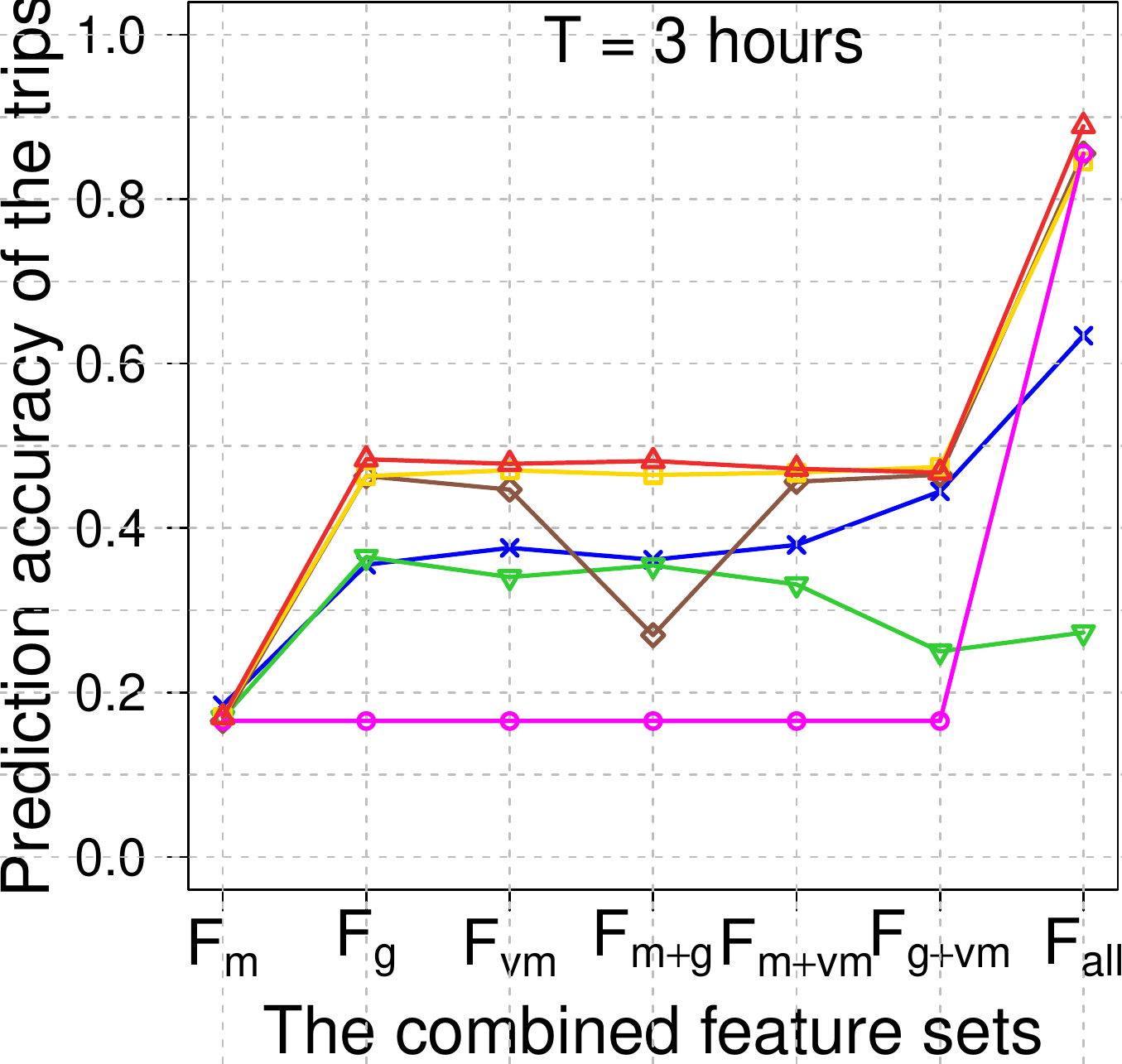}}
	\hspace{0in}
	\subfigure[Predicting the \emph{BTD} on the bus routes in the next 4 hours.]{
		\label{pic:accuracyH4} 
		\includegraphics[width=1.6in]{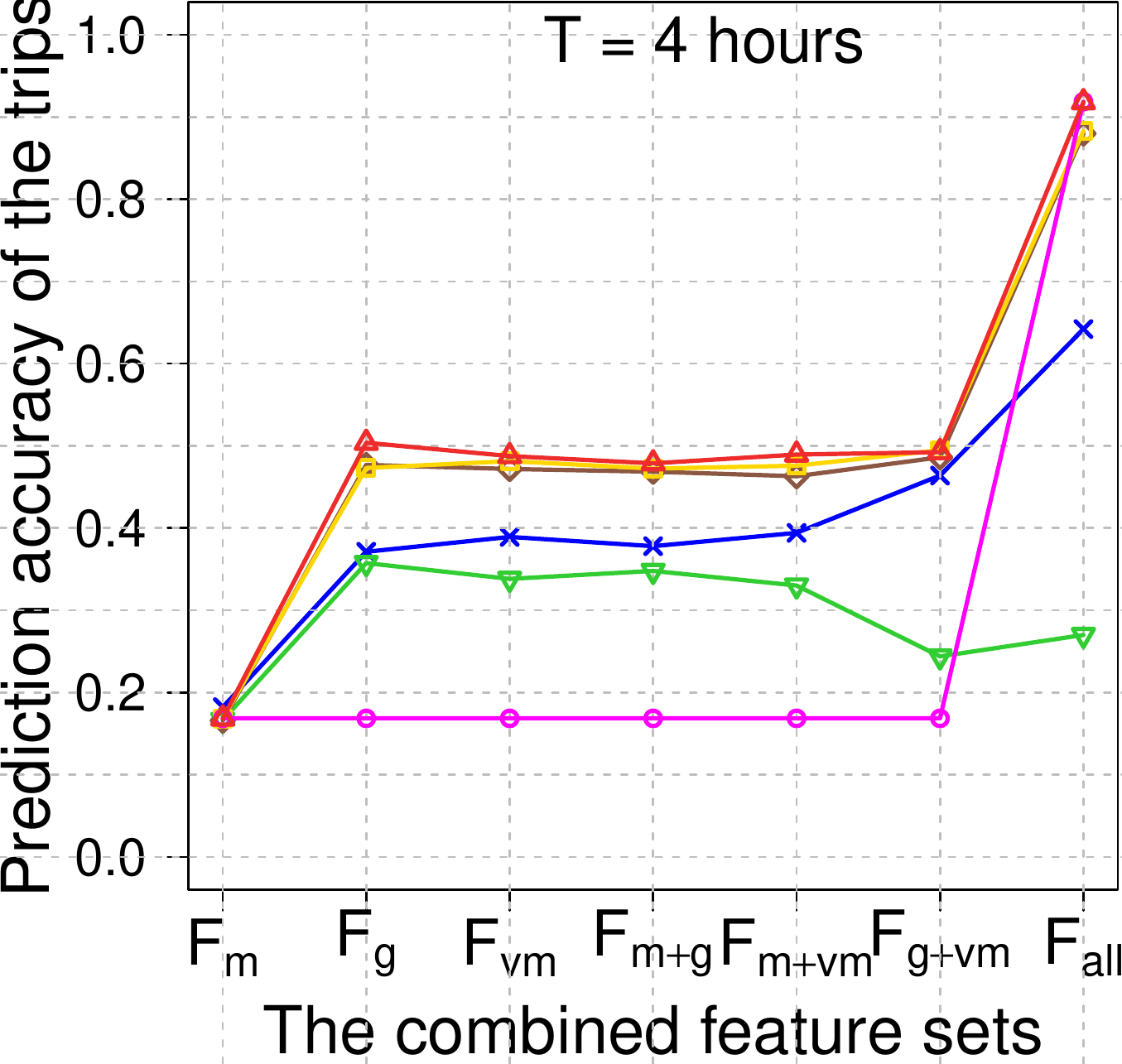}}
	\subfigure[Predicting the \emph{BTD} on the bus routes in the next 5 hours.]{
		\label{pic:accuracyH5}
		\includegraphics[width=1.6in]{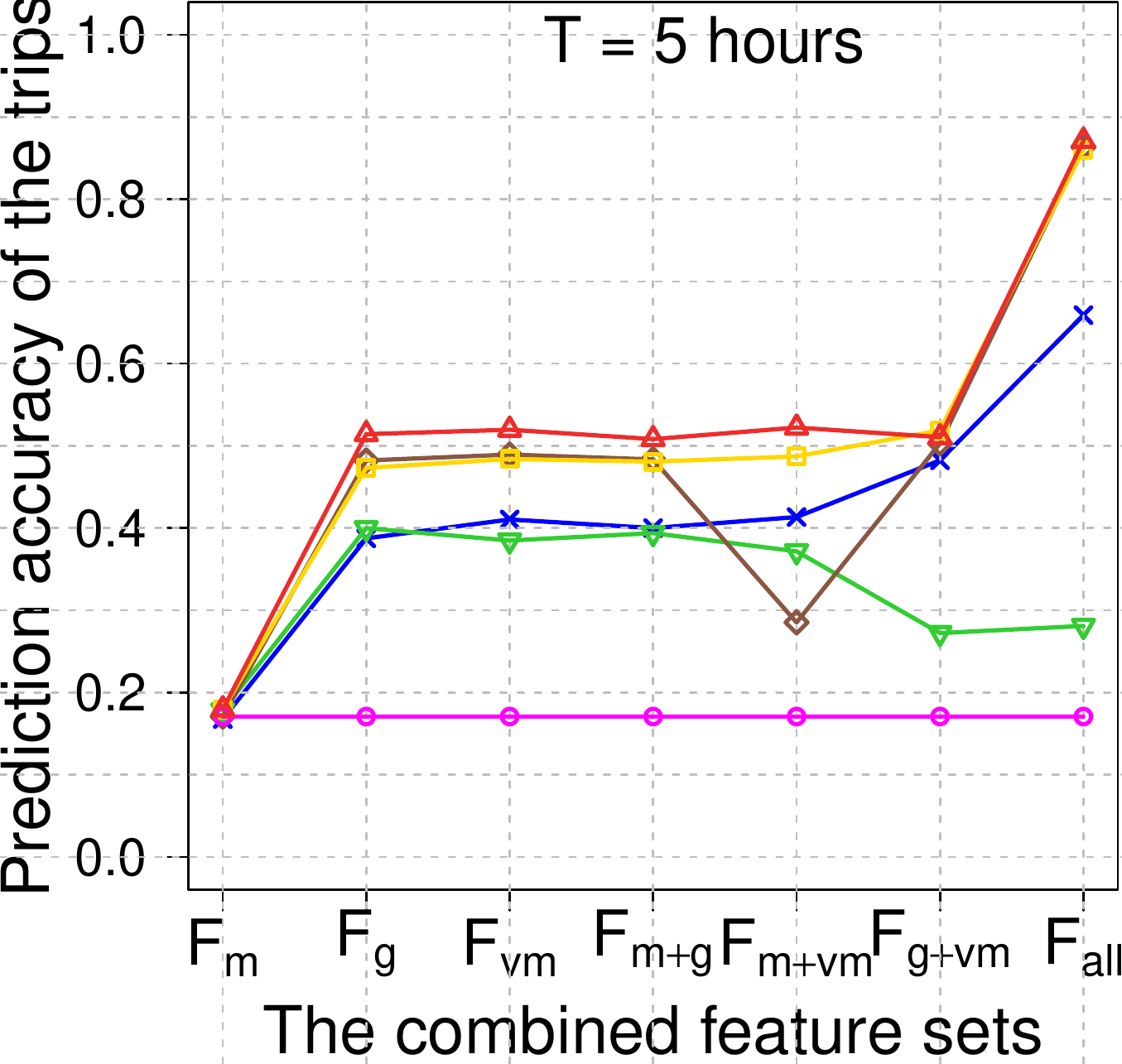}}
	\hspace{0in}
	\subfigure[Predicting the \emph{BTD} on the bus routes in the next 6 hours.]{
		\label{pic:accuracyH6} 
		\includegraphics[width=1.6in]{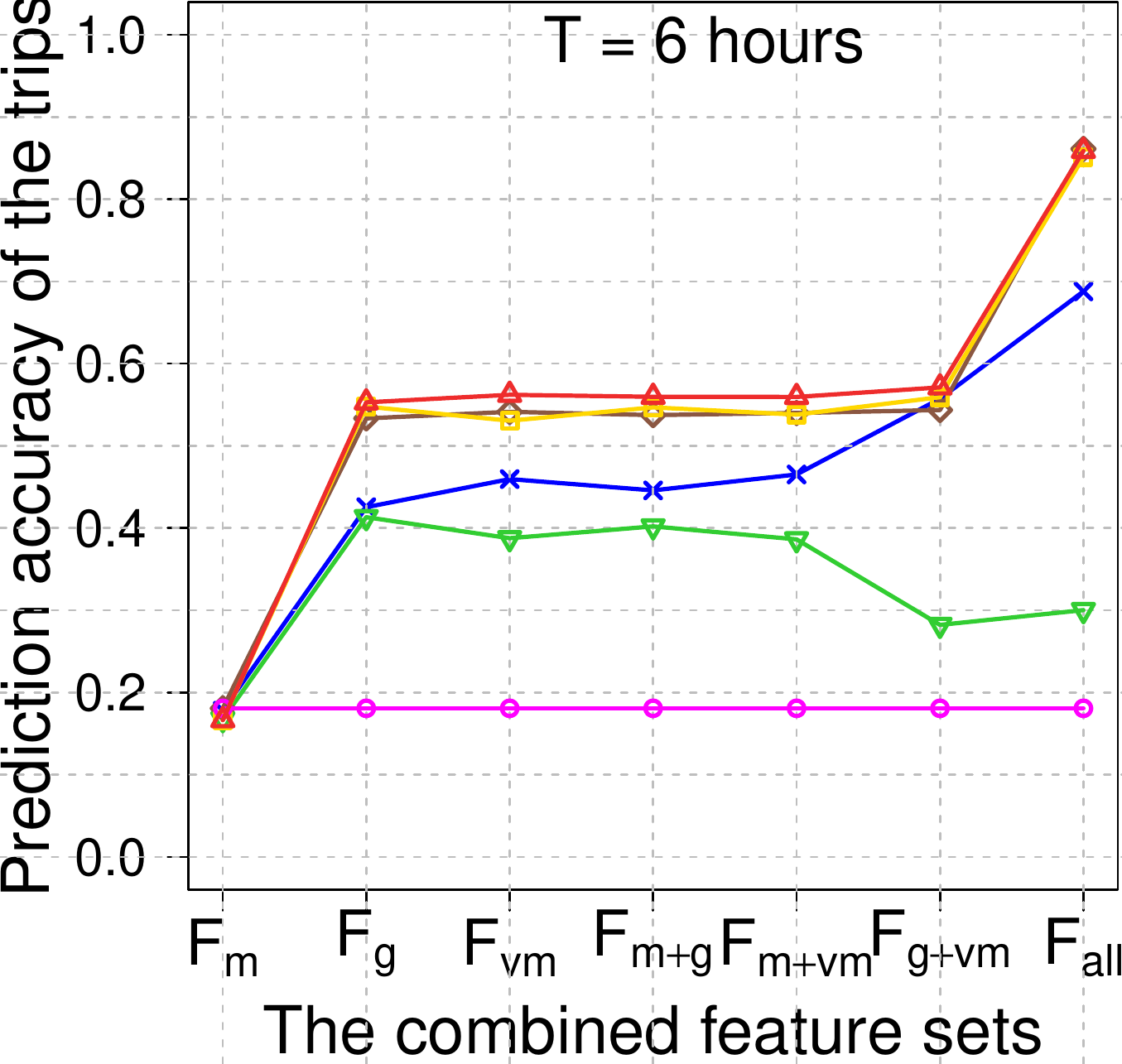}}
	\subfigure[Each algorithm is represented by the unique line and color.]{
		\label{pic:accuracyLag} 
		\includegraphics[width=3.4in]{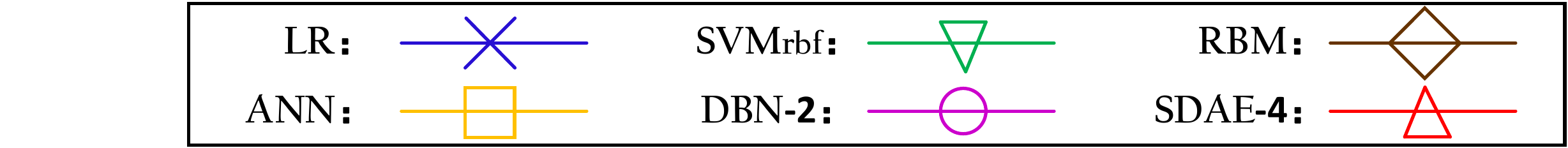}}
	\caption{Prediction accuracy of \emph{LR}, \emph{SVM}$_{rbf}$, \emph{RBM}, \emph{ANN}, \emph{DBN}-2, and \emph{SDAE}-4 with the diverse feature sets at different time intervals $T$. }
	\label{pic:accuracy} 
\end{figure}

In the experiments, we first justify the effectiveness of the extracted features for the \emph{Trip Demand Predictor}. Specifically, through the combination of  $F_{vm}$, $F_g$, and $F_{m}$ in different ways, we obtain seven kinds of feature sets, that is  $F_{vm}$, $F_g$, $F_{m}$, $F_{m+g}$, $F_{m+vm}$, $F_{g+vm}$, and $F_{all}$. $F_{m+g}$, $F_{m+vm}$, $F_{g+vm}$, and $F_{all}$ respectively are the fusions of $F_m$ and $F_g$, $F_m$ and $F_{vm}$, $F_g$ and $F_{vm}$, and all the feature sets. Furthermore, utilizing the proposed \emph{SDAE-4} algorithm and other baseline methods, we experiment with the seven feature sets to predict the number of people's bus trips (\emph{BTD}) at different periods in the future. The prediction accuracy of these methods is displayed in the Fig. \ref{pic:accuracy}.

In this figure, the first six sub-figures describe the prediction accuracy of people's trips on the bus routes in the next one to six hours. In the first six sub-figures, the x-axis represents the seven feature sets. The y-axis indicates the prediction accuracy of \emph{BTD} in the future period of time $T$. As shown in the sub-figure \ref{pic:accuracyLag}, each prediction method is  represented by the unique line and color. The \emph{DBN}-2 is a two-hidden-layer \emph{DBN}. Given the results of different methods, the prediction accuracy of people's bus trips varies dramatically with the change of feature sets. By fusing all the $F_{m}$, $F_{g}$, and $F_{vm}$, we can always achieve a dramatically improvement on the accuracy of the \emph{SDAE}-4 and other baselines at the different time intervals $T$. Moreover, whether based on individual feature set or their combinations, our proposed \emph{SDAE}-4 outperforms other baselines at each future time interval in most instances. Consequently, these studies further demonstrate that the feature sets we extracted are considerably useful for predicting \emph{BTD} and are noticeably effective for improving the prediction accuracy.

\noindent $\mathbf{2}$)\quad\noindent\textbf{\emph{Evaluating with different number of layers and different number of hidden units per layer:}}

\begin{figure}[ht]
	\subfigure[Predicting the \emph{BTD} on the bus routes in the next 1 hour.]{
		\label{pic:SDAEParaVary}
		\includegraphics[width=1.6in]{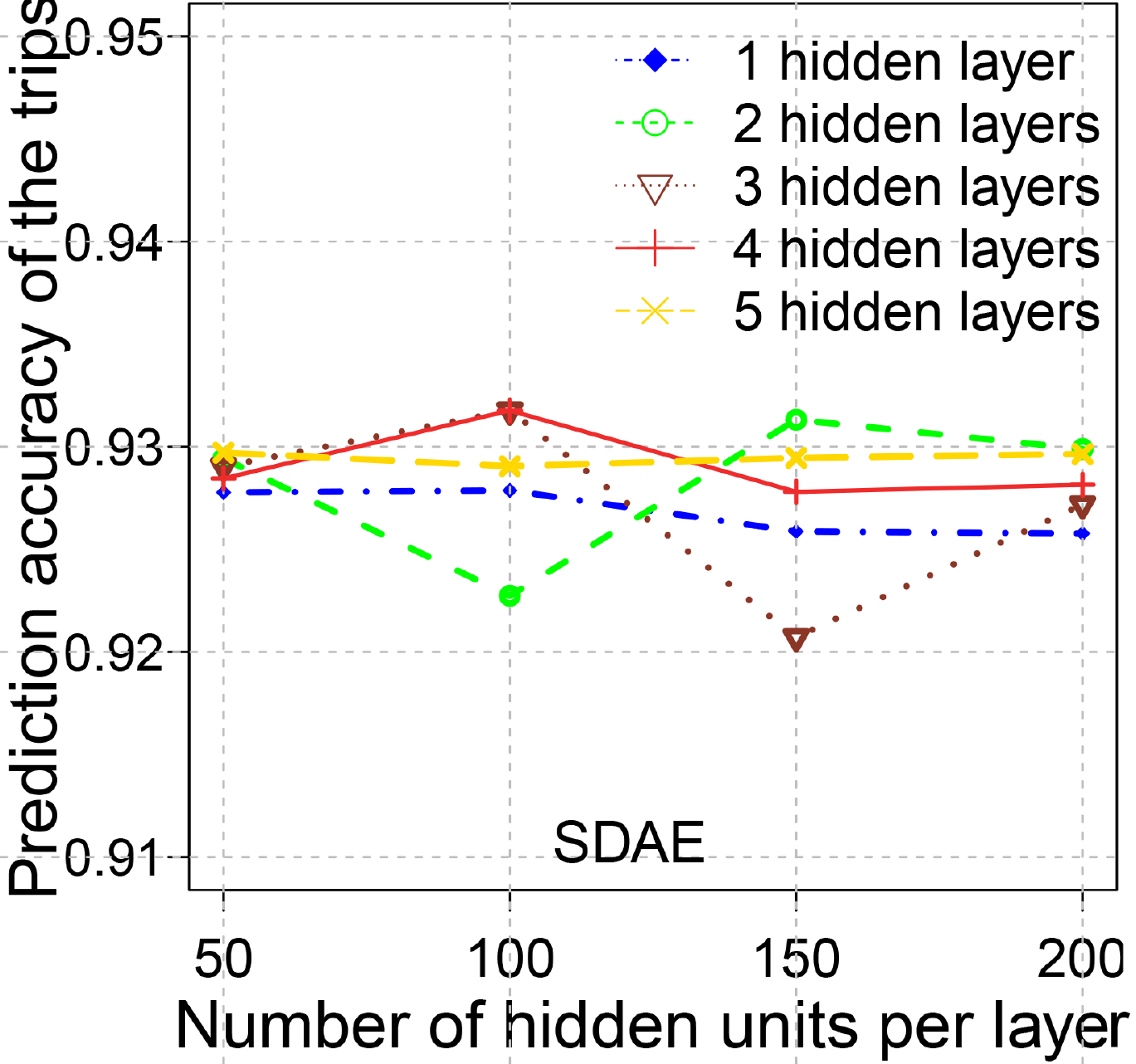}}
	\hspace{0in}
	\subfigure[Predicting the \emph{BTD} on the bus routes in the next 1 hour.]{
		\label{pic:DBNParaVary} 
		\includegraphics[width=1.6in]{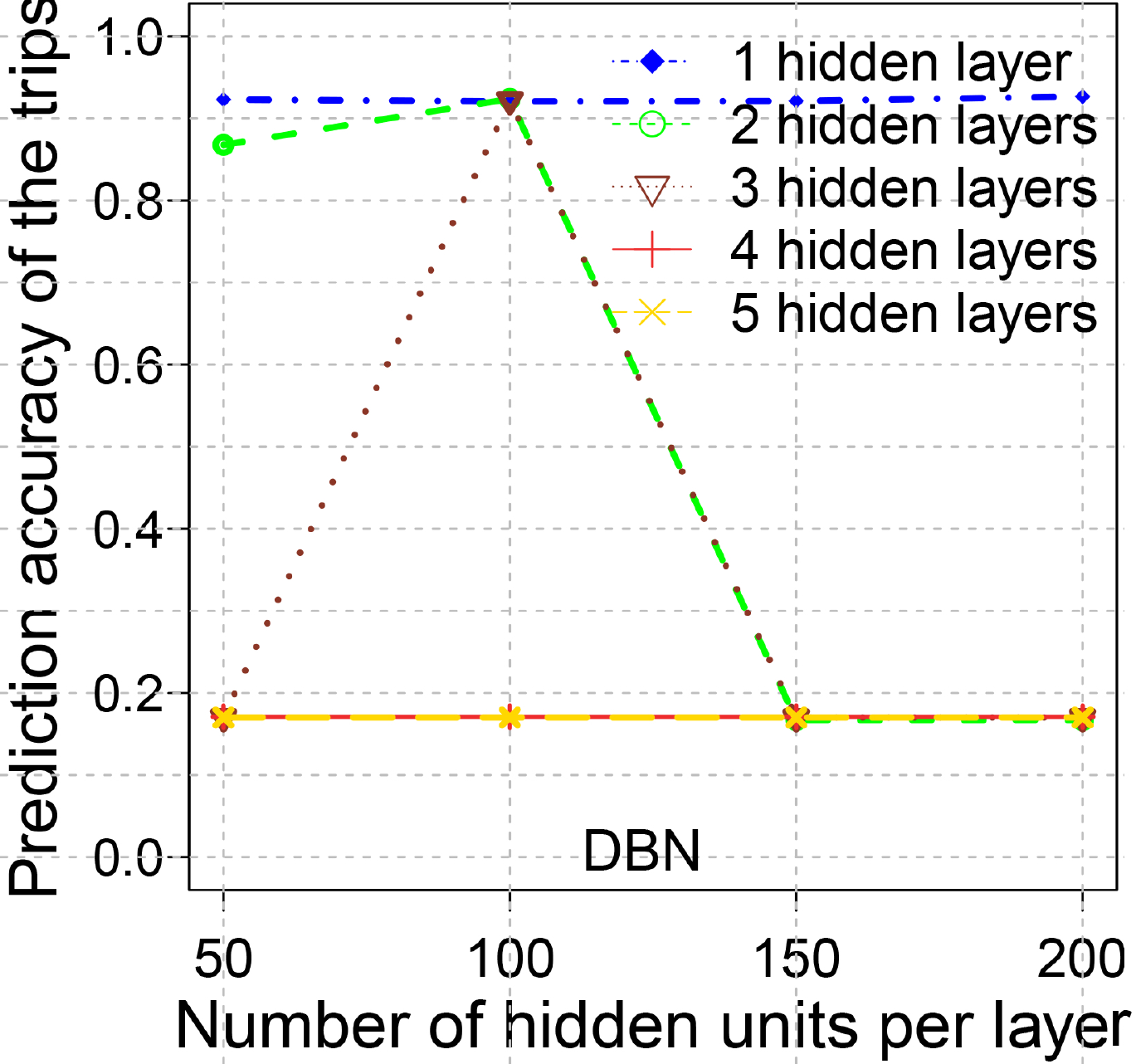}}
	\caption{Prediction performance on all feature sets for the deep neural network pre-trained by \emph{DBN} and \emph{SDAE}, as we increase the number of hidden layers and the number of neurons per layer.}
	\label{pic:EvaNumlayersUnits} 
\end{figure}

In this section, we mainly study the influence of the number of layers and the number of hidden units per layer on the prediction accuracy. More specifically, we aim to examine the behaviors of the \emph{SDAE} and the baseline \emph{DBN} on all the feature sets, as modifying the number of hidden layers and the number of neurons per layer. Fig. \ref{pic:EvaNumlayersUnits} displays the variation in the prediction accuracy of  \emph{SDAE} and \emph{DBN}, as we increase the number of hidden layers from 1 to 5 and the number of hidden units per layer from 50 to 200.  With the change of conditions, the prediction accuracy of the \emph{DBN} varies extremely, while the prediction accuracy of the \emph{SDAE} is relatively stable. Furthermore, compared with \emph{DBN}, the advantages of \emph{SDAE} become even more obvious with the increase of number of layers and number of hidden units. Additionally, for all the feature sets, the \emph{SDAE} with 4 hidden layers and 100 neurons per layer appears to perform better than other \emph{SDAE} variants and all \emph{DBN} variants. 

\noindent $\mathbf{3}$)\quad\noindent\textbf{\emph{Evaluating with different time spans:}}

\begin{figure}[ht]
	\centering
	\includegraphics[width=1.0\linewidth]{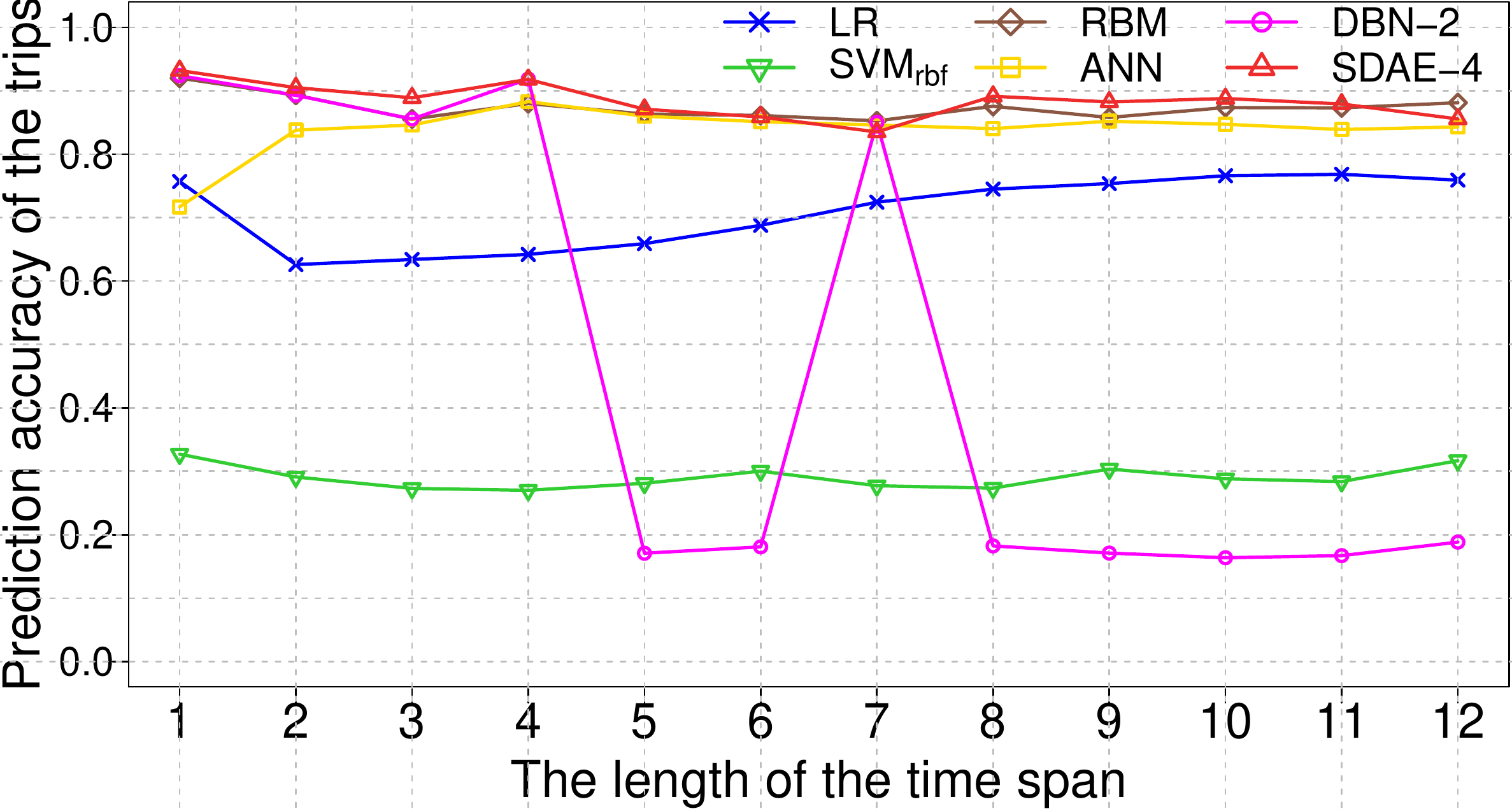}
	\caption{Prediction accuracy of the \emph{BTD} on all feature sets for \emph{SDAE-4} and the baselines, as we change the time span.}
	\label{pic:EvaTimeSpans}
\end{figure}

In previous experiments, the feature sets related to people's bus trips are extracted during a specified period of time. Here we evaluate how the length of time span affects the prediction performance of the proposed \emph{SDAE}-4 and other baselines. In the experiments, based on all the corresponding feature sets, we change the time span from $1$ to $12$ hours and measure the prediction precision of the proposed \emph{SDAE}-4 and other baselines for each time span, as displayed in Fig. \ref{pic:EvaTimeSpans}. The results indicate that the \emph{SDAE}-4, \emph{DBN}-2, \emph{RBM}, \emph{SVM}$_{rbf}$ and \emph{LR} all can obtain better prediction accuracy of people's bus trips for the next one hour. In addition, for all the methods except \emph{ANN}, in general their prediction accuracy will becomes poor with the increase of time span.
Moreover, in most cases, our proposed \emph{SDAE}-4 is always superior to other baselines for all the time spans from $1$ to $12$ hours and has a stronger robustness to predict people's bus trips in the future.

\vspace{0.18cm}
\noindent $\mathbf{4}$)\quad\noindent\textbf{\emph{Evaluation on route planning efficiency:}}

In this section, we mainly study the efficiency of our route planning system in reducing carbon emissions. Our experiments are conducted in Zhuhai, where the proportion of electric buses is relatively small in all buses and each electric bus is allocated to a fixed route. Moreover, for traditional gasoline buses, the energy consumption is about 38 liters per 100 kilometers, and the carbon dioxide emission coefficient is about 2.73 kilograms per liters. Based on the experimental results, the efficiency of our designed system is mainly manifested in two aspects: 1) cutting the peak of carbon emissions in the city; 2) improving the utilization rate of electric buses. Given the route recommendation mechanism for electric buses, the recommended routes  always cross the place where people's trip need will be considerably large and the amount of transportation carbon emissions would reach the higher levels. In the place that any electric bus goes through, the peak of carbon emissions will be reduced by $0.90l$, where the $l$ is the transport mileage traveled by the electric bus across each area.

\noindent $\mathbf{5}$)\quad\noindent\textbf{\emph{Overall results:}}

\begin{figure}[ht]
	\centering
	\includegraphics[width=0.8\linewidth]{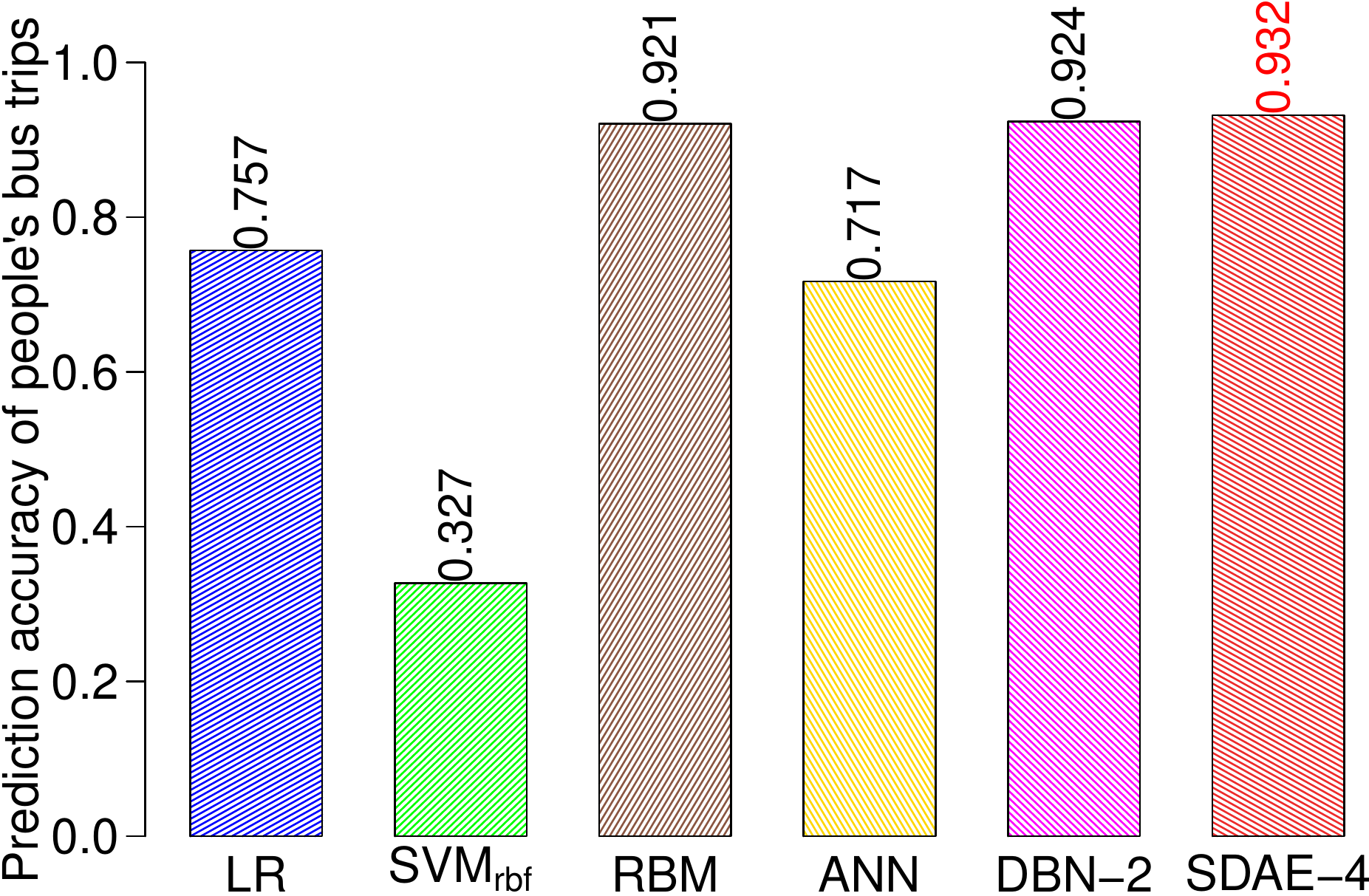}
	\caption{Prediction accuracy of our proposed \emph{SDAE}-4 and all the baselines on all feature sets for the next 1 hour.}
	\label{pic:OverallResults}
\end{figure}

In our study the experimental data are  divided into three parts: training data of $234,192$ records extracted from 82 days,  validation data of $28,560$ records extracted from 10 days, and test data  $39,984$ records extracted from 14 days. Meanwhile, given the combined feature sets $F_{all}$, we utilize the proposed \emph{SDAE}-4 and five baselines to predict the \emph{BTD} on the bus routes in the next one hour. Specifically, our proposed \emph{SDAE}-4 has four hidden layers, and each hidden layer has 100 neurons. The baseline \emph{DBN}-2 has two hidden layers, and each hidden layer has 100 neurons. 
As displayed in Fig. \ref{pic:OverallResults}, \emph{SDAE}-4 consistently outperforms the baselines. The prediction accuracy of \emph{SDAE}-4 being above 0.932. 
Furthermore, in the experiments, the last layer of \emph{SDAE}-4 and \emph{DBN}-2 are both the \emph{LR} layers. Compared with the \emph{SDAE}-4 and \emph{DBN}-2, the prediction accuracy of the \emph{LR} method in Fig. \ref{pic:OverallResults} is relatively poor, that being about 0.757. Accordingly, the prediction performance of \emph{LR} in last layer of \emph{SDAE}-4 and \emph{DBN}-2 benefit significantly from using the higher level representations. These higher level representations learnt by stacking denoising autoencoders or by stacking restricted Boltzmann machines significantly improve the performance of \emph{SDAE}-4 and \emph{DBN}-2.

\begin{figure}[ht]
	\subfigure[ Distribution of the \emph{BTD} on the routes in an hour after 11 a.m. every day in September 2015.]{
		\label{pic:BTDVaries30Days}
		\includegraphics[width=1.60in]{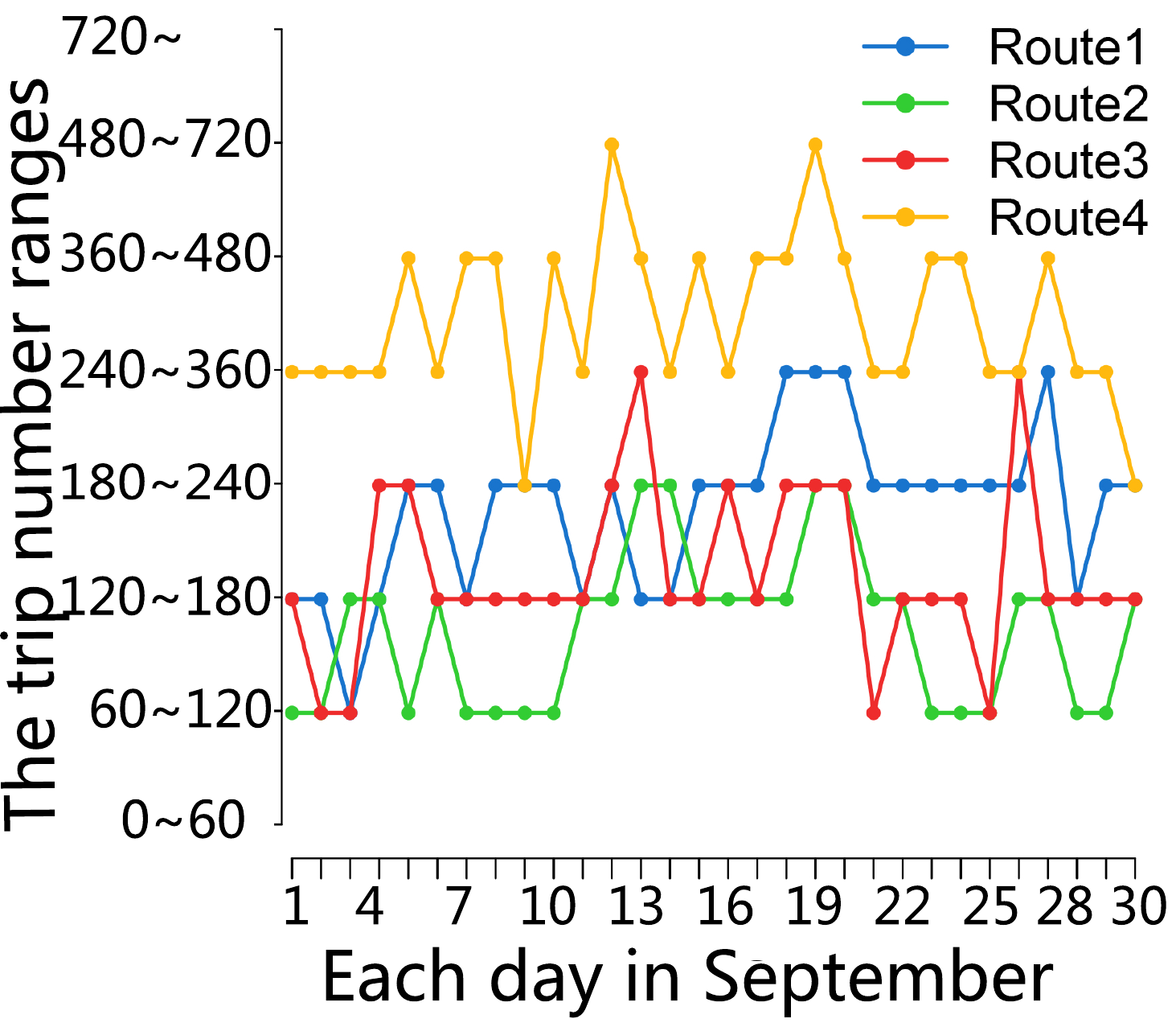}}
	\hspace{0in}
	\subfigure[ Distribution of the \emph{BTD} on the different routes in an hour of September 19, 2015.]{
		\label{pic:BTDVaries10Days5Hours} 
		\includegraphics[width=1.6in]{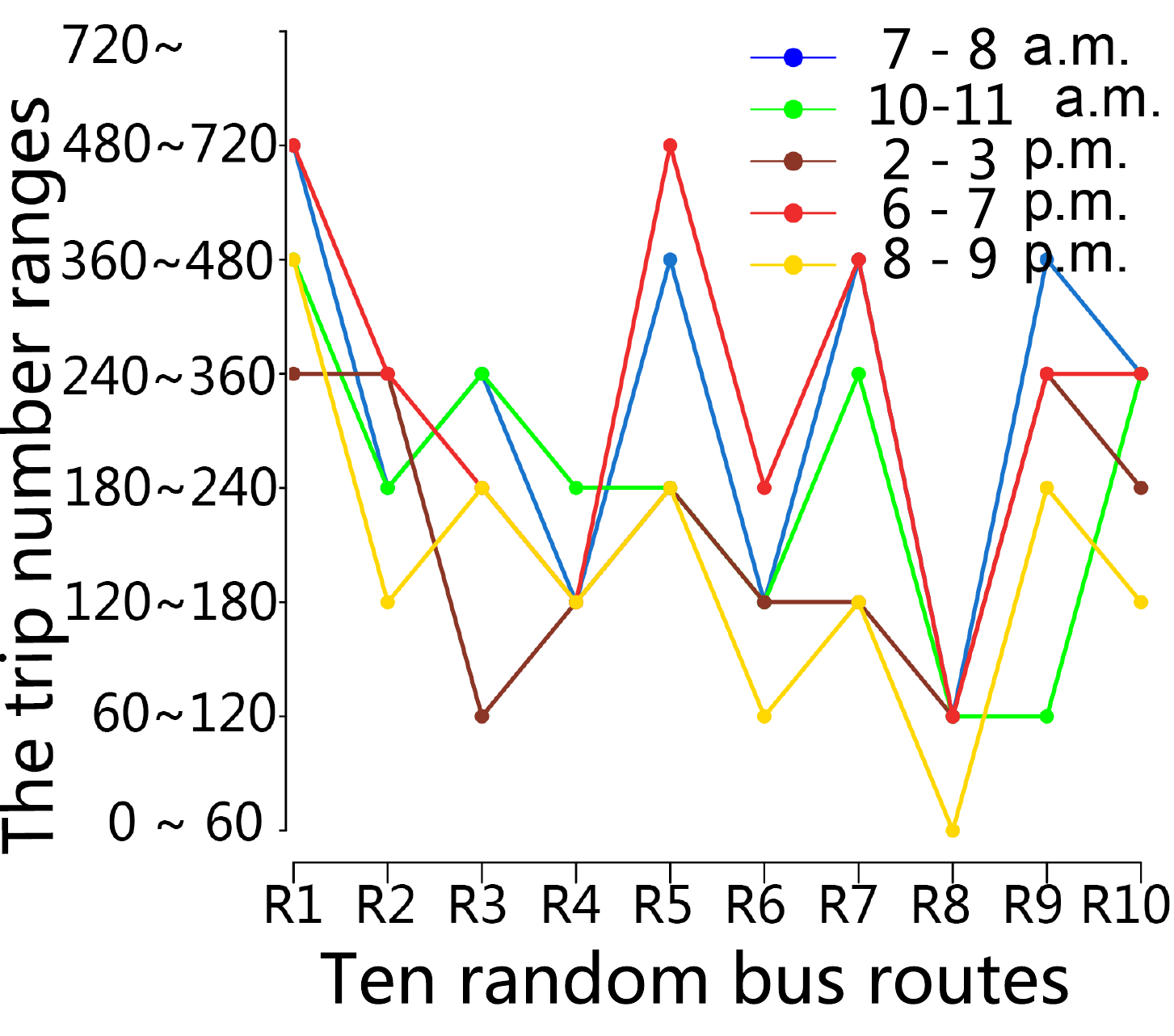}}
	\caption{ Distribution of the \emph{BTD} predicted by the proposed \emph{SDAE}-4 on all feature sets.}
	\label{pic:EvaBTDVaries} 
\end{figure}

Moreover, given the proposed \emph{SDAE}-4, Fig. \ref{pic:EvaBTDVaries} clearly illustrates the distribution of the predicted \emph{BTD} in two case: on the same route in an hour after 11 a.m. every day in September 2015 and on the different routes in the same hour of September 19, 2015. 
For any bus route in Fig. \ref{pic:BTDVaries30Days} such as \emph{Route1, Route2, Route3 or Route4}, the \emph{BTD} changes  noticeably in the hour after 11 a.m. every day in September. Meanwhile, as shown in Fig. \ref{pic:BTDVaries10Days5Hours},  even in the same period of a day, the predicted \emph{BTD} are always varies with the bus route. For instance, during the period between 6 p.m. and 7 p.m., the distributions of \emph{BTD} are considerably diverse on the selected ten routes (\emph{R1, R2, $\cdots$, R10}). The results indicate that the proposed \emph{SDAE}-4 can well learn the specific spatial-temporal characteristics of each bus route and accurately predict the \emph{BTD} under local conditions. 

\section{Conclusion}
In this paper, we devise a \emph{Clairvoyance} system to solve a route planning problem for electric buses that will depart in an ideal state, based on the extracted spatial-temporal and semantic features about vehicle movement, human mobility, geographic location, and weather. Specifically, the proposed  \emph{Clairvoyance} leverages the \emph{SDAE-4} network to predict the future number of people's bus trips on each bus route and utilizes our three-layer perceptron network\cite{lu2017predicting} to forecast the future transportation carbon emission in each region throughout the city. Furthermore, given the future information of people's bus trips and carbon emissions, a greedy strategy is introduced to recommend top-\emph{k} bus routes for each electric bus that will depart. Our experiments are carried out on the real-world urban big data in Zhuhai, China.
The results indicate that the extracted features are considerably effective for improving the performance of the proposed two networks. Depending on these features, our networks both outperform the other well-known baselines, such as \emph{SVM}, \emph{RBN}, \emph{ANN} and \emph{DBN}. Moreover, we also quantitatively evaluate the efficiency of route planning in terms of reducing carbon emissions.

\emph{Clairvoyance} provides a novel deep learning-based strategy to assign suitable routes for electric buses, based on the heterogeneous urban big data. The available urban datasets play a crucial role in our system. Accordingly, the limitation of \emph{Clairvoyance} is that 1) the extracted features are from limited available urban datasets; 2) the \emph{Clairvoyance} does not consider the charging problem of electric buses. It is mainly based on a prior hypothesis that each electric bus that will depart is in an ideal battery state.
In the future, we will first test the robustness of our system on diverse datasets extracted from different cities and study which datasets determine our system's performance. Second, this paper does not consider the impact of the electric bus charging problem, which needs further study. Third, we will explore the impact of urban emergency events on electric bus dispatching, in order to provide the necessary emergency strategy to our system.


%



\section*{Acknowledgment}

The work is supported by NSFC (No.61472149), the Fundamental Research Funds for the Central Universities (2015QN67) and the National 863 Hi-Tech Research and Development Program under grant (2015AA01A203), JSPS KAKENHI Grant Number 16K00117, 15K15976 and KDDI Foundation.

\ifCLASSOPTIONcaptionsoff
  \newpage
\fi

\bibliographystyle{IEEEtran}
\bibliography{electricBus}

\begin{IEEEbiography}[{\includegraphics[width=1in, height=1.25in, clip, keepaspectratio]{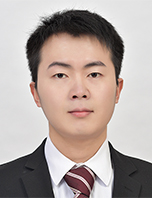}}]{Xiangyong Lu} is currently pursuing the Ph.D. degree in the Graduate School of Information Sciences at Tohoku University, Japan, under the guidance of Prof. Takayuki Okatani. He received the B.S. degree in computer science from Henan University, Kaifeng, China, in 2015. He received the M.S. degree in the Services Computing Technology and System Lab, Big Data Technology and System Lab, Cluster and Grid Computing Lab, School of Computer Science and Technology, Huazhong University of Science and Technology, Wuhan, 430074, China, under the guidance of Prof. Chen Yu. From April 2016 to March 2017, he was a visiting scholar in the Emerging Networks and Systems Lab and Wireless Networks Lab, the Department of Information and Electronic Engineering, Muroran Institute of Technology, Muroran, Hokkaido, Japan, under the guidance of Prof. Kaoru Ota. His research interests include computer vision, deep learning, ubiquitous computing and data mining. 
\end{IEEEbiography}

\vspace{-1.2cm}
\begin{IEEEbiography}[{\includegraphics[width=1in, height=1.25in, clip, keepaspectratio]{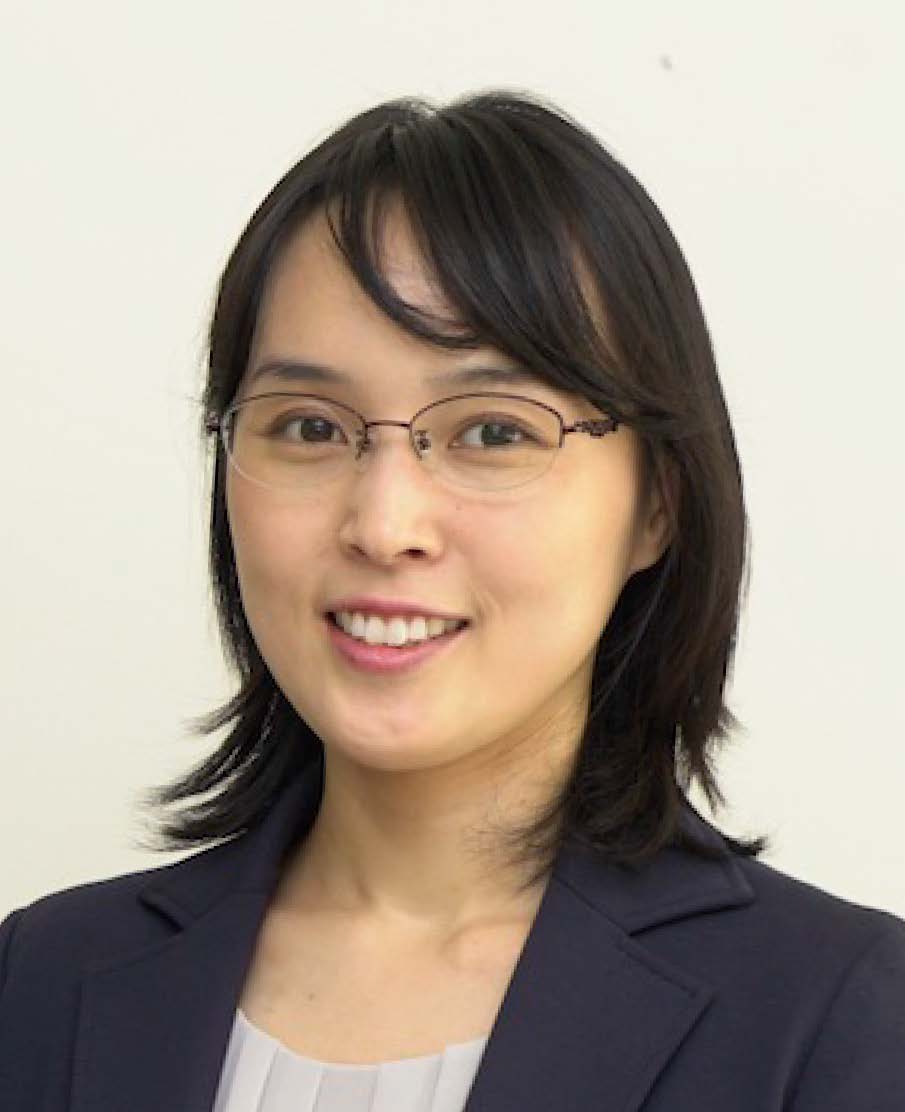}}]{Kaoru Ota} was born in Aizu-Wakamatsu, Japan. She received M.S. degree in Computer Science from Oklahoma State University, the USA in 2008, B.S. and Ph.D. degrees in Computer Science and Engineering from The University of Aizu, Japan in 2006, 2012, respectively. Kaoru is currently an Associate Professor and Ministry of Education, Culture, Sports, Science and Technology (MEXT) Excellent Young Researcher with the Department of Sciences and Informatics, Muroran Institute of Technology, Japan. From March 2010 to March 2011, she was a visiting scholar at the University of Waterloo, Canada. Also, she was a Japan Society of the Promotion of Science (JSPS) research fellow at Tohoku University, Japan from April 2012 to April 2013. Kaoru is the recipient of IEEE TCSC Early Career Award 2017, and The 13th IEEE ComSoc Asia-Pacific Young Researcher Award 2018. She is Clarivate Analytics 2019 Highly Cited Researcher (Web of Science).
\end{IEEEbiography}


\vspace{-1.2cm}

\begin{IEEEbiography}[{\includegraphics[width=1in, height=1.25in, clip, keepaspectratio]{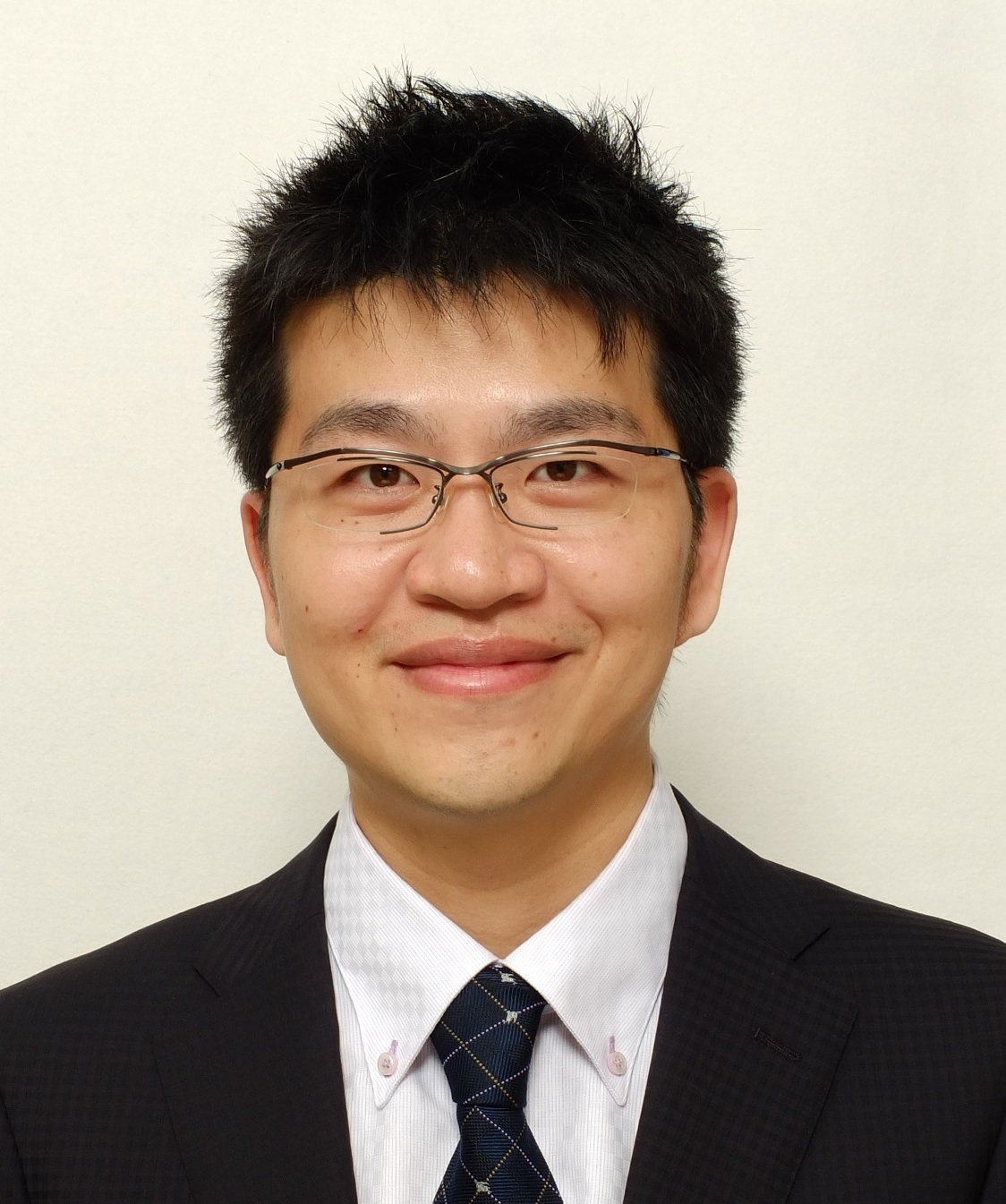}}]{Mianxiong Dong} received B.S., M.S. and Ph.D. in Computer Science and Engineering from The University of Aizu, Japan. He is the Vice President and youngest ever Professor of Muroran Institute of Technology, Japan. He was a JSPS Research Fellow with School of Computer Science and Engineering, The University of Aizu, Japan and was a visiting scholar with BBCR group at the University of Waterloo, Canada supported by JSPS Excellent Young Researcher Overseas Visit Program from April 2010 to August 2011. Dr. Dong was selected as a Foreigner Research Fellow (a total of 3 recipients all over Japan) by NEC C$\&$C Foundation in 2011. He is the recipient of IEEE TCSC Early Career Award 2016, IEEE SCSTC Outstanding Young Researcher Award 2017, The 12th IEEE ComSoc Asia-Pacific Young Researcher Award 2017, Funai Research Award 2018 and NISTEP Researcher 2018 (one of only 11 people in Japan) in recognition of significant contributions in science and technology. He is Clarivate Analytics 2019 Highly Cited Researcher (Web of Science). 
\end{IEEEbiography}



\vspace{-1.2cm}

\begin{IEEEbiography}[{\includegraphics[width=1in, height=1.25in, clip, keepaspectratio]{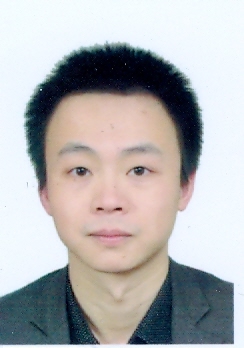}}]{Chen Yu} received the B.S. degree in mathematics and the M.S. degree in computer science from Wuhan University, Wuhan, China, in 1998 and 2002, respectively, and the Ph.D. degree in information science from Tohoku University, Sendai, Japan, in 2005.
	
	From 2005 to 2006, he was a Japan Science and Technology Agency Postdoctoral Researcher with the Japan Advanced Institute of Science and Technology. In 2006, he was a Japan Society for the Promotion of Science Postdoctoral Fellow with the Japan Advanced Institute of Science and Technology. Since 2008, he has been with the School of Computer Science and Technology, Huazhong University of Science and Technology, Wuhan, where he is currently a full Professor and a Special Research Fellow, working in the areas of wireless sensor networks, ubiquitous computing, and green communications. Dr. Yu was a recipient of the Best Paper Award in the 2005 IEEE International Conference on Communication and the nominated Best Paper Award in the Proceedings of the 11th IEEE International Symposium on Distributed Simulation and Real-Time Application in 2007. 
\end{IEEEbiography}

\vspace{-1.2cm}

\begin{IEEEbiography}[{\includegraphics[width=1in, height=1.25in, clip, keepaspectratio]{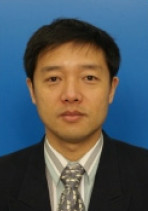}}]{Hai Jin}	is a Cheung Kung Scholars Chair Professor of computer science and engineering at Huazhong University of Science and Technology (HUST) in China. Jin received his PhD in computer engineering from HUST in 1994. In 1996, he was awarded a German Academic Exchange Service fellowship to visit the Technical University of Chemnitz in Germany. Jin worked at The University of Hong Kong between 1998 and 2000, and as a visiting scholar at the University of Southern California between 1999 and 2000. He was awarded Excellent Youth Award from the National Science Foundation of China in 2001. Jin is the chief scientist of ChinaGrid, the largest grid computing project in China, and the chief scientists of National 973 Basic Research Program Project of Virtualization Technology of Computing System, and Cloud Security. 
	
	Jin is an IEEE Fellow and a member of the ACM. He has co-authored 22 books and published over 700 research papers. His research interests include computer architecture, virtualization technology, cluster computing and cloud computing, peer-to-peer computing, network storage, and network security.
\end{IEEEbiography}








\end{document}